\documentclass{article} 
\usepackage{iclr2025_conference,times}

\usepackage{amsmath,amsfonts,bm}

\def\eqref#1{equation~\ref{#1}}

\def\1{\bm{1}}

\DeclareMathAlphabet{\mathsfit}{\encodingdefault}{\sfdefault}{m}{sl}
\SetMathAlphabet{\mathsfit}{bold}{\encodingdefault}{\sfdefault}{bx}{n}

\iclrfinalcopy 

\usepackage{hyperref}
\usepackage{graphicx}
\usepackage{url}
\usepackage{natbib}
\usepackage{xcolor}
\usepackage{booktabs}
\usepackage{wrapfig}
\usepackage{comment}
\usepackage{tcolorbox}

\newcommand{\opportunity}[1]{\textbf{\textcolor{red}{POTENTIAL OPPORTUNITY FOR TECHNICAL DEPTH\\}}}

\usepackage{xspace}
\newcommand{\ClaudeSonnet}{{Claude 3.5 Sonnet}\xspace}  
\newcommand{\ClaudeOpus}{{Claude 3 Opus}\xspace}  
\newcommand{\GeminiPro}{{Gemini 1.5 Pro}\xspace}  

\newcommand{\GPTFourO}{{GPT-4o 2024-05-13}\xspace}  
 
\newcommand{\GPTFourTurboApril}{{GPT-4 Turbo 2024-04-09}\xspace} 

\usepackage{amsthm}

\title{Unearthing Skill-Level Insights for Understanding Trade-Offs of Foundation Models}

\author{Mazda Moayeri$^{1,2}$\thanks{Work done while interning at Microsoft Research}\;, Vidhisha Balachandran$^1$\thanks{equal contribution}\;, Varun Chandrasekaran$^{1,3\dagger}$, \AND Safoora Yousefi$^1$, Thomas Fel$^4$, Soheil Feizi$^2$, Besmira Nushi$^1$, Neel Joshi$^1$\thanks{co-prinicipal investigators}\;, Vibhav Vineet$^{1\ddagger}$ \\
\\
$^1$Microsoft Research AI Frontiers, $^2$University of Maryland, \\$^3$University of Illinois Urbana-Champaign, $^4$Harvard University Kempner Institute\\
\texttt{mmoayeri@umd.edu, vivineet@microsoft.com} \\
}

\begin{document}

\maketitle

\begin{abstract}


With models getting stronger, evaluations have grown more complex, testing multiple skills in one benchmark and even in the same instance at once. However, skill-wise performance is obscured when inspecting aggregate accuracy, under-utilizing the rich signal modern benchmarks contain. We propose an automatic approach to recover the underlying skills relevant for any evaluation instance, by way of inspecting model-generated {\em rationales}. After validating the relevance of rationale-parsed skills and inferring skills for $46$k instances over $12$ benchmarks, we observe many skills to be common across benchmarks, resulting in the curation of hundreds of \emph{skill-slices} (i.e. sets of instances testing a common skill). Inspecting accuracy over these slices yields novel insights on model trade-offs: e.g., compared to GPT-4o and \ClaudeSonnet, on average, \GeminiPro is $18\%$ more accurate in \emph{computing molar mass}, but $19\%$ less accurate in \emph{applying constitutional law}, despite the overall accuracies of the three models differing by a mere $0.4\%$. Furthermore, we demonstrate the practical utility of our approach by showing that insights derived from skill slice analysis can generalize to held-out instances: when routing each instance to the model strongest on the relevant skills, we see a $3\%$ accuracy improvement over our $12$ dataset corpus. 
Our skill-slices and framework open a new avenue in evaluation, leveraging skill-specific analyses to unlock a more granular and actionable understanding of model capabilities. 



\end{abstract}

\section{Introduction}

\begin{figure}
    \centering
    \includegraphics[width=\linewidth]{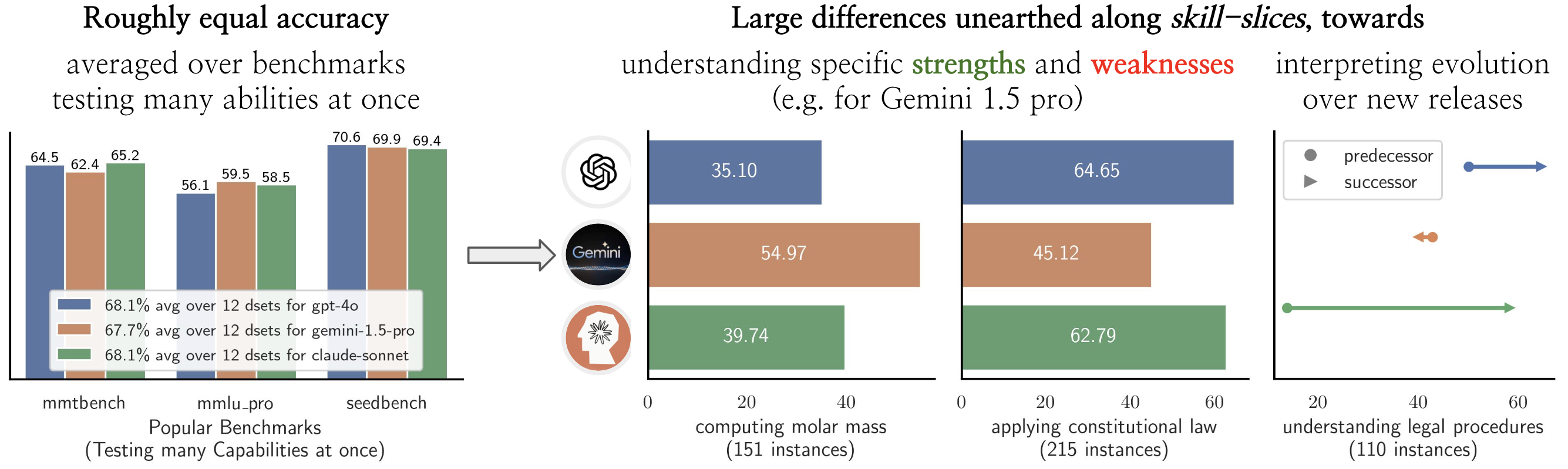}
    \caption{We leverage model-generated rationales to extract the skills relevant to any evaluation instance. Inspecting accuracy along \emph{skill-slices} (instances drawn across benchmarks involving the same skill) surfaces fine-grained insights otherwise obfuscated by aggregate accuracy.}
    \label{fig:teaser}
\end{figure}

Recent years have seen benchmarks evolve to keep up with ever-advancing models. While classical benchmarks tested specific capabilities, like recognizing digits \citep{lecun1998mnist} or classifying sentiment \citep{bowman2015snli}, modern benchmarks measure proficiency in numerous capabilities simultaneously, drawing questions of increasing difficulty from more diverse domains \citep{mialon2024gaia, yue2023mmmu, wang2024mmluprorobustchallengingmultitask}. As the questions we test models on have grown more complex, aggregate performance measures provide less understanding about model proficiency in specific abilites. 
For example, as shown in Figure \ref{fig:teaser}, we find that over a dozen benchmarks, GPT-4o, \GeminiPro, and \ClaudeSonnet \citep{openai2024gpt4o, geminiteam2024gemini15unlockingmultimodal, anthropic2024claude35sonnet} achieve overall accuracies within $0.4\%$ of one another, leaving an open question: Are these models all the same, or are valuable insights being averaged away?

\looseness=-1
Manual annotations of categories across instances (e.g. \cite{mmtbench} and \cite{liu2024mmbench} include `ability' tags) enable going beyond accuracy, but at a cost for benchmark creators that rises with the number and difficulty of test questions. This results in few (if any) and non-standardized annotations, restricting cross-benchmark aggregation, even though many benchmarks have large overlap in the (implicit) skills they test \citep{miao-etal-2020-diverse, cobbe2021gsm8k}. Prior works show promise in automatically grouping inputs by attributes to extract image classification failure modes (e.g. `fails on \emph{white} foxes') \citep{eyuboglu2022domino, rezaei2024prime}, but these methods fail to produce deeper insights on the underlying competencies that models use across contexts (e.g. `fails on \emph{counting}').

To this end, we propose a method to automatically discover \emph{skills} related to any evaluation instance, toward a finer-grained understanding of model capabilities from existing benchmarks. We consider skills as latent features pertaining to ``how'' a model must operate -- the steps involved in solving a given task -- whereas a general attribute relates to ``what'' is being addressed -- the observable characteristics of the instance. Moreover, the relevant skills for an instance may not be readily apparent through superficial examination, often only revealing themselves upon inspection of the solution. 
This complexity highlights the challenge of automatically inferring relevant skills and underscores the need to develop new approaches for their effective identification.

\looseness=-1
In this paper, we harness the \emph{rationale}-generating ability of strong models to infer skills, as rationales (i.e. explained solutions) elucidate the steps involved in performing a task, and have had many benefits in the past \citep{wei2022chain, kojima, Singh2024RethinkingII, xai_rationales, orca2, hsieh2023distilling}. Importantly, we use rationales not to directly analyze the rationale-generating model, but instead to annotate \emph{data}, on which \emph{any} model can be evaluated.



Specifically, given an evaluation instance, we instruct a strong model (e.g. GPT-4o) to generate a detailed step-by-step rationale, along with the skill applied in each step. We then parse skills per instance for numerous benchmarks and aggregate instances along \textbf{\emph{skill-slices}}, i.e. subsets of instances drawn from multiple benchmarks that all involve a specific skill, enabling measurement of model proficiency for that skill. After applying our technique to over $45k$ instances from a dozen popular modern benchmarks (primarily multimodal), we find \emph{hundreds} of skills with slice sizes of $\geq100$. To validate the relevance of our skills, we devise an automatic verification protocol and observe that our automatically extracted skills are relevant, even in cases where the annotating model fails in solving the query problem. We will release all rationales, skill annotations, and skill-slices, which we term the \emph{Skill-Index}, to the public at \textcolor{blue}{\url{github.com/microsoft/skill-slice-insights}}.

\looseness=-1
Having verified the accuracy of our constructed skill-slices, we present multiple ways our rationale-based skill annotations and skill-slices can be used to inform our understanding and usage of current large models. First, we conduct detailed skill-level analyses of 6 state-of-art models from the GPT, Gemini, and Claude families, \textbf{identifying practical insights into model strengths and weaknesses at previously underexplored granularities}. Despite roughly equal average accuracy across our corpus for GPT-4o, \GeminiPro, and \ClaudeSonnet, we discover skills for each model  where accuracy far exceeds that of its counterparts, with gaps as large as $20\%$ (see Figure \ref{fig:teaser}). For example, we find \GeminiPro to be much stronger in math and science skills, like `computing molar mass', while falling far behind for legal skills like `applying constitutional law'. Next, we show that our \textbf{skill-slices allow for fine-grained understanding of improvements in new models within a model family}: while \GeminiPro did not improve over v1.0 on `understanding legal procedures', OpenAI and Claude models improved substantially, with \ClaudeSonnet leaping nearly $50\%$ over \ClaudeOpus. 

Further, our \textbf{skill-level understanding of models can be directly utilized for instance level model selection} based on skills required. When routing each query instance to a model with best performance on relevant skills, we see improvements in overall accuracy across a dozen benchmarks by $3\%$, including gains of $3.5$ to $7\%$ on MMLU Pro \citep{wang2024mmluprorobustchallengingmultitask}. Finally, to assess a model's proficiency on a skill in isolation, we generate probing questions that directly target a specific skill. We find that model performance on the synthetic probing sets corroborates accuracies on our constructed skill-slices, further strengthening our findings.

\looseness=-1

\looseness=-1
Our skill based framework and findings advocate for a paradigm shift in foundation model evaluation, emphasizing the importance of skill-specific analysis to gain a more granular and actionable understanding of model capabilities. We summarize key contributions below:

\begin{itemize}

\item Automated Skill Inference: We present a scalable method that leverages model-generated rationales to recover (and validate) the underlying skills relevant to any evaluation instance.

\item Novel findings from \emph{Skill-Slices}: Analyzing skill-slices -- sets of instances testing a common skill -- reveals new fine-grained insights on the trade-offs across leading models. 

\item Skill Annotations for Popular Benchmarks: We release the \emph{Skill-Index}, a dataset of instance-level skill annotations and rationales, along with code to easily expand to new benchmarks, providing a valuable resource for the research community.

\end{itemize}
\section{Constructing skill-slices using Model-Generated Rationales}
\label{sec:method}
We now detail our method for inferring relevant skills for each evaluation instance and aggregating skill annotations across datasets to form \emph{skill-slices} (i.e. a set of instances that share a relevant skill). We utilize \emph{rationales}, or step-by-step solutions, generated by a strong model (e.g. GPT-4o) to facilitate the extraction of relevant skills.
However, as evaluation instances are challenging by design and generating relevant skills requires a significant level of meta-reasoning, it is important to assess the quality of produced skills.
Thus, we additionally present a multifaceted automatic approach to validating the relevance of extracted skills, which we confirm aligns well with human judgments.

\begin{figure}
    \centering
    \includegraphics[width=0.98\linewidth]{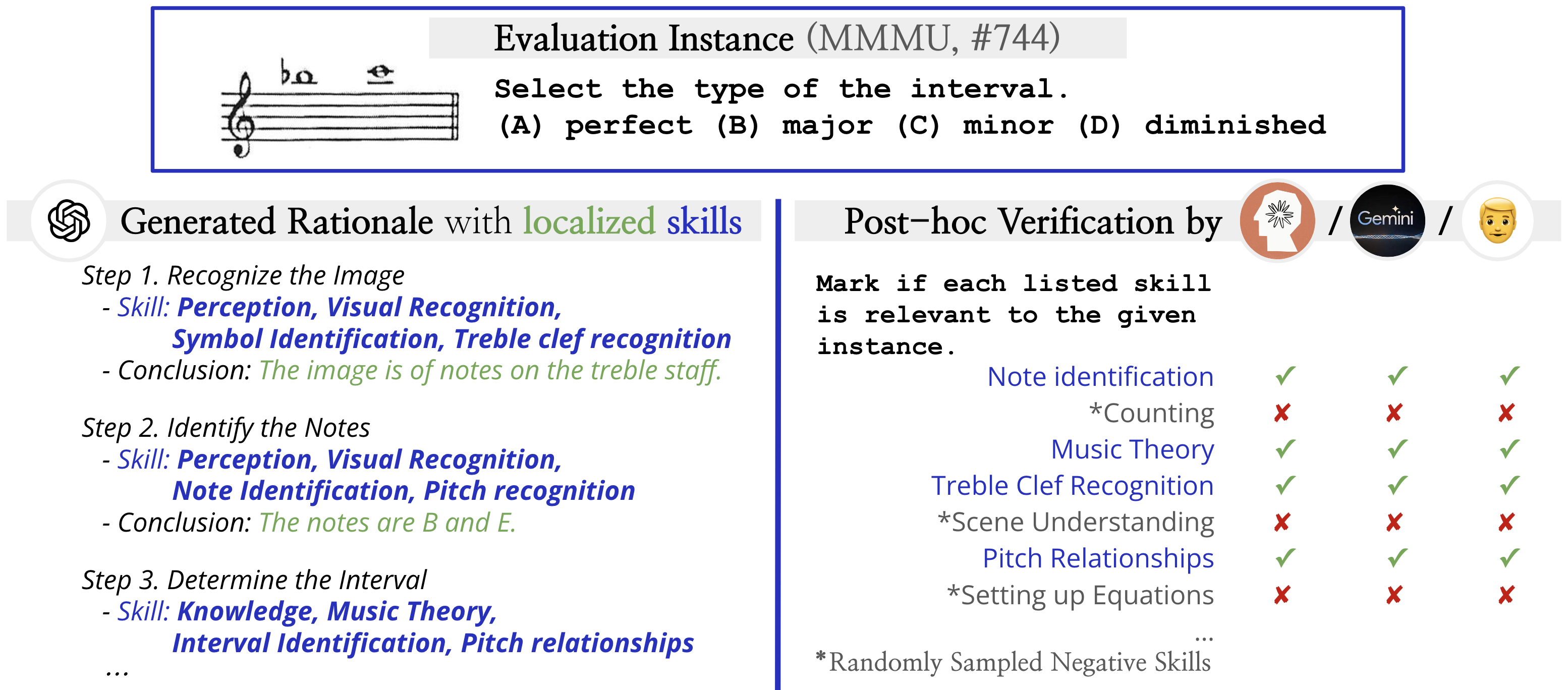}
    \caption{\textbf{(left)} Sample GPT-4o generated rationale: each skill is listed under multiple names of \emph{cascaded granularity}, and localized to a specific step and concluding claim. \textbf{(right)} Annotated skills can be verified independently with a second model or human. We include randomly sampled negative skills and multiple verifiers to assure the quality of the verification, as detailed in \ref{subsec:verifiers}.}
    \label{fig:methods_diagrams}
\end{figure}

\subsection{Skill Extraction via Rationale Parsing}
\label{subsec:skill-extraction}
Given an evaluation instance, we prompt GPT-4o to generate a detailed rationale (i.e. step-by-step solution), where each step only involves the use of a single skill; figure \ref{fig:methods_diagrams} presents an example. We additionally instruct the model to, after each step, list the applied skill with \emph{multiple} names of \emph{cascading granularity}. For example, the skill `treble clef recognition' is prepended with `perception, visual recognition, symbol identification'\footnote{If some of these skills are unfamiliar to you, you're not alone! Many benchmarks today ask questions that are challenging to anyone without expertise in the relevant domain. This precisely motivates relying on stronger models to help us make sense of these benchmarks, as manual inspection has increasingly limited utility.}. Listing multiple names of different granularity per skill is a simple way to increase the number of slices an evaluation instance falls into, leading to a greater quantity and diversity of skill-slices. Further, each granularity has unique and complementary advantages: fine-grained skills are more specific, while coarse-grained skills lead to slices with a larger number of samples, making an accuracy estimate over such slices more reliable. Also, in order to standardize the response structure, we include an in-context example in our prompt, which enables simple parsing of rationales to extract skills. See Appendix \ref{app-sec:prompt} for complete details.

\looseness=-1
Note that other methods could possibly be used to annotate relevant skills, like directly prompting GPT-4o to list them. We prioritized maximizing the total count and size of resultant skill-slices. In comparing the skills per instance obtained by direct prompting vs. our rationale parsing, we indeed find that rationale parsing results in a higher average count and diversity in grain (Appendix \ref{app-sec:prompt-ablation}). 


\subsection{Curating Skill-Slices across Benchmarks}

\looseness=-1
With rationale-parsing, we tag relevant skills for $46$k evaluation instances from 12 benchmarks\footnote{See Appendix \ref{app-sec:skill-set} for complete details on datasets annotated and the slice formation process.} that are popular (i.e. commonly featured in evaluations for recent model releases) and unsaturated (SOTA under $90\%$ accuracy), resulting in $690$k total skills ($128$k unique) over $202$k rationale steps. Importantly, we observe that \textbf{skills cut across benchmarks}. That is, many skill-slices of non-trivial size are formed when including instances across benchmarks, because \emph{models re-use skills in diverse contexts}. Namely, $278$ skill-slices have at least $100$ unique instances each.
This number rises to $332$ after de-duplicating via a tight clustering\footnote{We employ a high minimum cosine similarity threshold of $0.95$ per cluster to de-duplicate skills listed under slightly different names; e.g., `trigonometry' and `trigonometric calculation' have a similarity of $0.95$.} on the text embeddings of the skills. Aggregating across benchmarks enables us to analyze skills that are in the long tail for one benchmark, but get resurfaced when many different benchmarks are joined together. Indeed, slice count increases by $51\%$ when slicing across benchmarks instead of only drawing instances from one benchmark at a time.

\begin{figure}
    \centering
    \begin{minipage}{0.36\linewidth}
    \includegraphics[width=\linewidth]{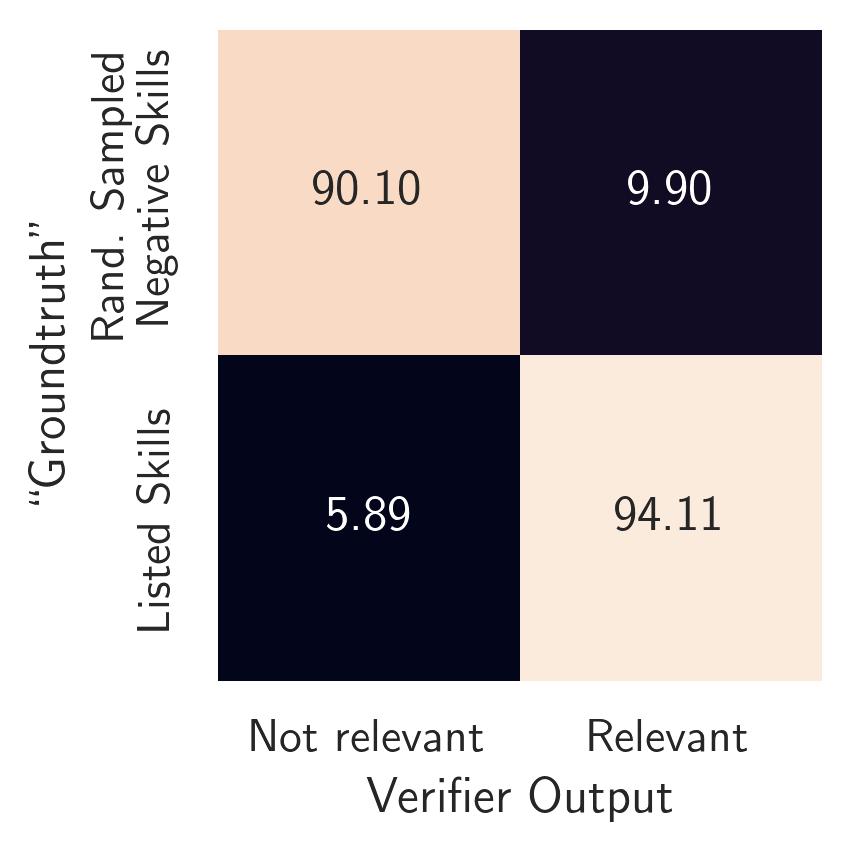}
    \end{minipage}
    \begin{minipage}{0.45\linewidth}
    \includegraphics[width=\linewidth]{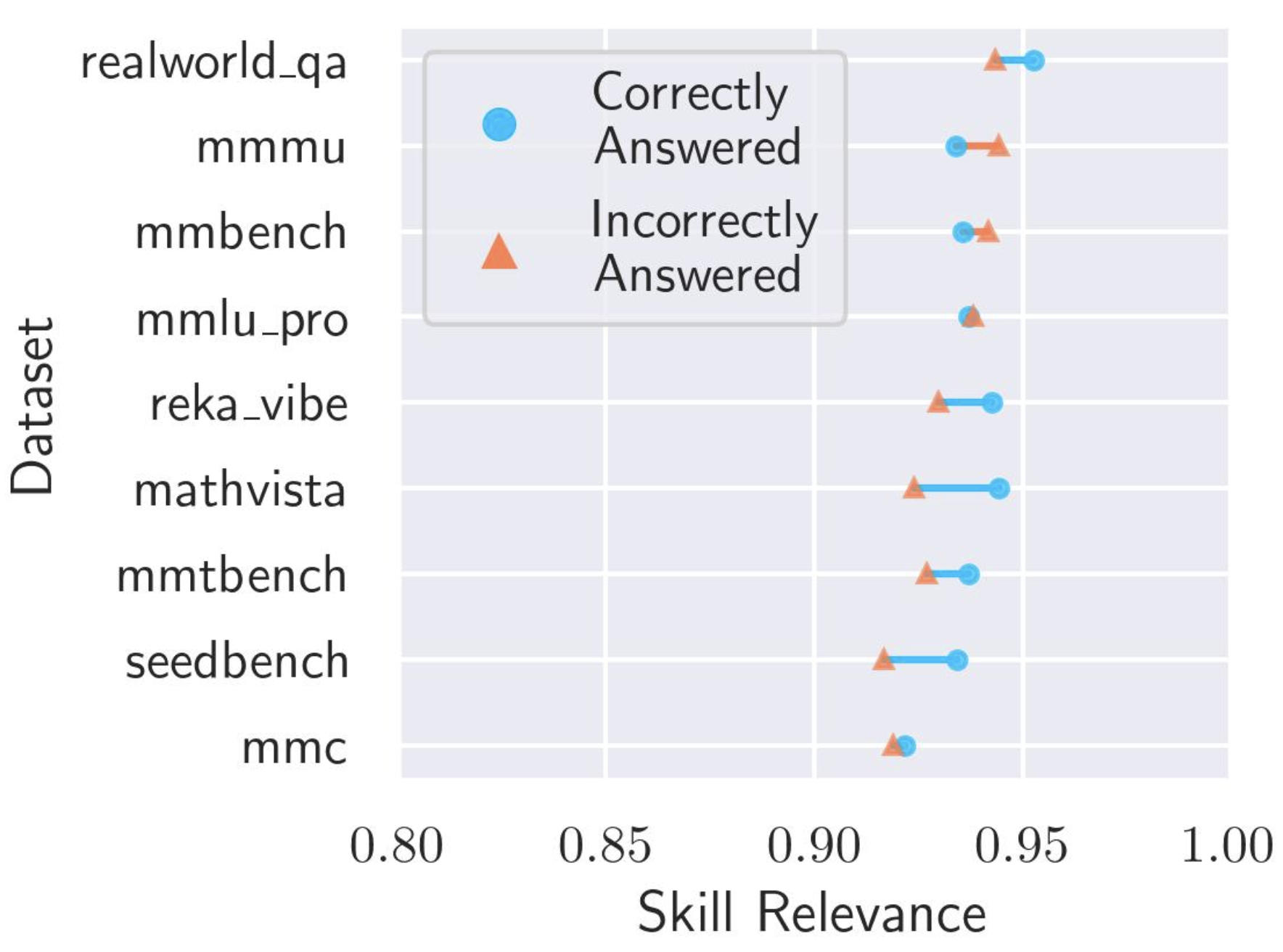}
    \end{minipage}
    \caption{(\textbf{left}) Post-hoc verification shows GPT-4o-annotated skills are relevant, and that automatic verifiers are reliable, as they admit low rates of false positives (marking a randomly sampled negative skill as relevant). (\textbf{right}) Rate that GPT-4o-annotated skills are marked as relevant, separated by if GPT-4o correctly answered the underlying evaluation instance (blue) or not (orange). Empirically, annotated skills have high relevancy rates \emph{even} when the annotator incorrectly answers the question.}
    \label{fig:validation}
\end{figure}

\subsection{Automated Validation of Skill Relevance}
\label{subsec:verifiers}
\looseness=-1

\looseness=-1
In the absence of fine-grained ground truth skill annotations that would confirm whether generated skills are relevant or not, we design an automatic validation approach.
Namely, we propose two automated methods to directly validate the relevancy of listed skills: \emph{post-hoc verification} and \emph{inter-(skill)annotator agreement}. The latter checks the overlap in skills listed for the same instance by two different annotators. Here, we focus on the former and defer full details of both approaches to Appendix \ref{app-sec:validation}. Post-hoc verification consists of a second `verifier' model which inspects an evaluation instance with a list of skills and marks whether each skill is relevant or not.
We form the list of skills by combining the skills for the instance with an equally sized list of \emph{negative} skills, randomly sampled as follows: with $S$ denoting all skills from our corpus, negative skills for an instance $x$ with skills $S_x$ are drawn from $\{s \in S \mid \max_{s_i \in S_x}\text{sim}(s, s_i) \leq \tau\}$. Here, `sim' is the similarity of two skills, computed via the cosine similarity of text embeddings from a frozen text embedder. The threshold $\tau$ is selected based on the text embedder so that no negative skill is equivalent up to paraphrasing to an annotated skill for the instance. We include negative skills as quality controls to ensure that the verifier model does not simply mark all skills as relevant.

\looseness=-1
Figure \ref{fig:validation} shows the results of post-hoc verification of skills for $100$ skills per dataset averaged over three verifier models (\ClaudeSonnet, \GeminiPro, GPT-4v\footnote{GPT-4v denotes \GPTFourTurboApril, and GPT-4o denotes \GPTFourO throughout our paper. Also, \ClaudeSonnet refers to the model released on 2024-06-20, not the later revision of 2024-10-22.}). We observe $94.1\%$ of skills annotated by GPT-4o are verified as relevant. A smaller scale human validation of $640$ total skills results in $95.7\%$ relevancy rate and $92.8\%$ agreement with the automatic verifiers, indicating that post-hoc verification is a reliable automated metric for assessing skill relevancy. 
The right panel shows that the relevancy rate is roughly equivalent, \emph{regardless of if the annotator correctly answers the underlying evaluation instance}; crucially, this enables using automated skill annotations on instances that are interesting to evaluate on (i.e. hard enough to stump some models). 
Thus, while rationales may at times be partially incorrect\footnote{In fact, there's an error in the rationale in Figure \ref{fig:methods_diagrams}: the notes are $B\flat$ \& $C$. However, the skills are accurate.}, they can still be leveraged to shed insight on \emph{data} at scale.

\section{Analyzing Foundation  Models with Skill-slices}
\label{sec:findings}
\begin{figure}
    \centering
    \includegraphics[width=1\linewidth]{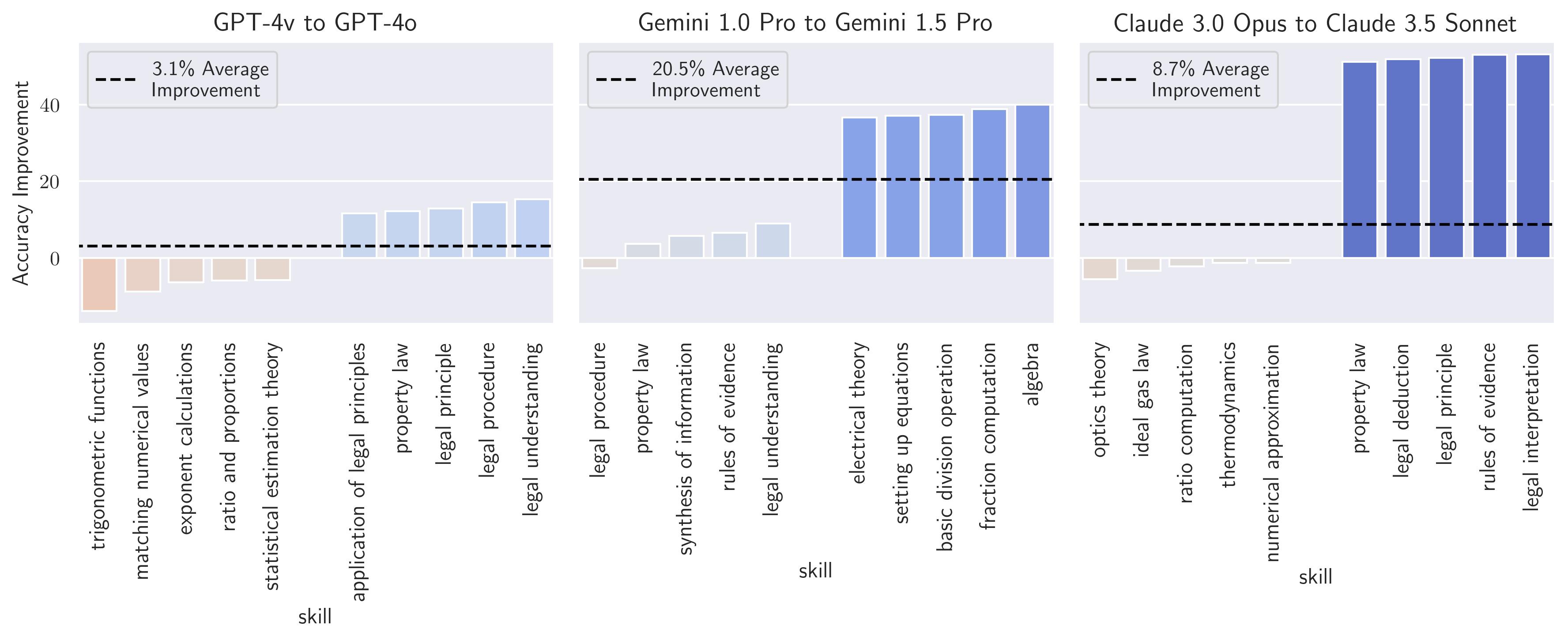}
    \caption{Skill-slices shed insight on how models evolve over new releases. For GPT and Claude models, skills related to law see the largest increases in accuracy, while Gemini models improve most in performing math and science skills.}
    \label{fig:model_evolution}
\end{figure}

\looseness=-1
We now leverage the skill-slices obtained in the previous section to better understand and utilize frontier models from OpenAI, Google, and Anthropic.
%
First, we identify skill-slices with vastly different accuracies across releases from the same family, as well as across families, which are obfuscated when inspecting overall accuracy. Then, having discovered models differ in their specific strengths and weaknesses, we show how overall accuracy can be improved by choosing the best model per instance or dataset, leveraging inferred skills and computed model-wise skill accuracies.




\subsection{Finer-grained Insights from Existing Evaluation Instances}

\looseness=-1
\textbf{Understanding Model Evolution.} 
Figure \ref{fig:model_evolution} shows average improvement, along with skill-slices with greatest and least improvement, for the last two releases in the Open AI GPT, Google Gemini, and Anthropic Claude families\footnote{Because Gemini-1.0-pro is not multimodal, we present results for skill-slices curated from $12k$ language-only evaluation instances (instead of our entire corpus) with at least $100$ instances within each slice.}. For all model families, we find skills where improvement is more than twice the average, as well as skills where the model does not improve at all. Notably, both GPT and Claude models see greatest increases in skills related to law, with Claude-Sonnnet 3.5 improving over Claude-Opus 3.0 by a staggering $\sim 50\%$ along numerous law skill-slices, suggesting improving legal abilities of their models may have been a recent priority for OpenAI and Anthropic. 
\begin{tcolorbox}[width=\textwidth,boxsep=0.01mm] 
   Finding 1: \textbf{Skill-level improvements across model releases greatly differ (i) from average improvement and (ii) between families.} The biggest leaps for GPT and Claude families were for legal skills, while Gemini improved most for math and science skills (figure \ref{fig:model_evolution}). 
\end{tcolorbox} 


\begin{figure}
    \centering
    \includegraphics[width=1\linewidth]{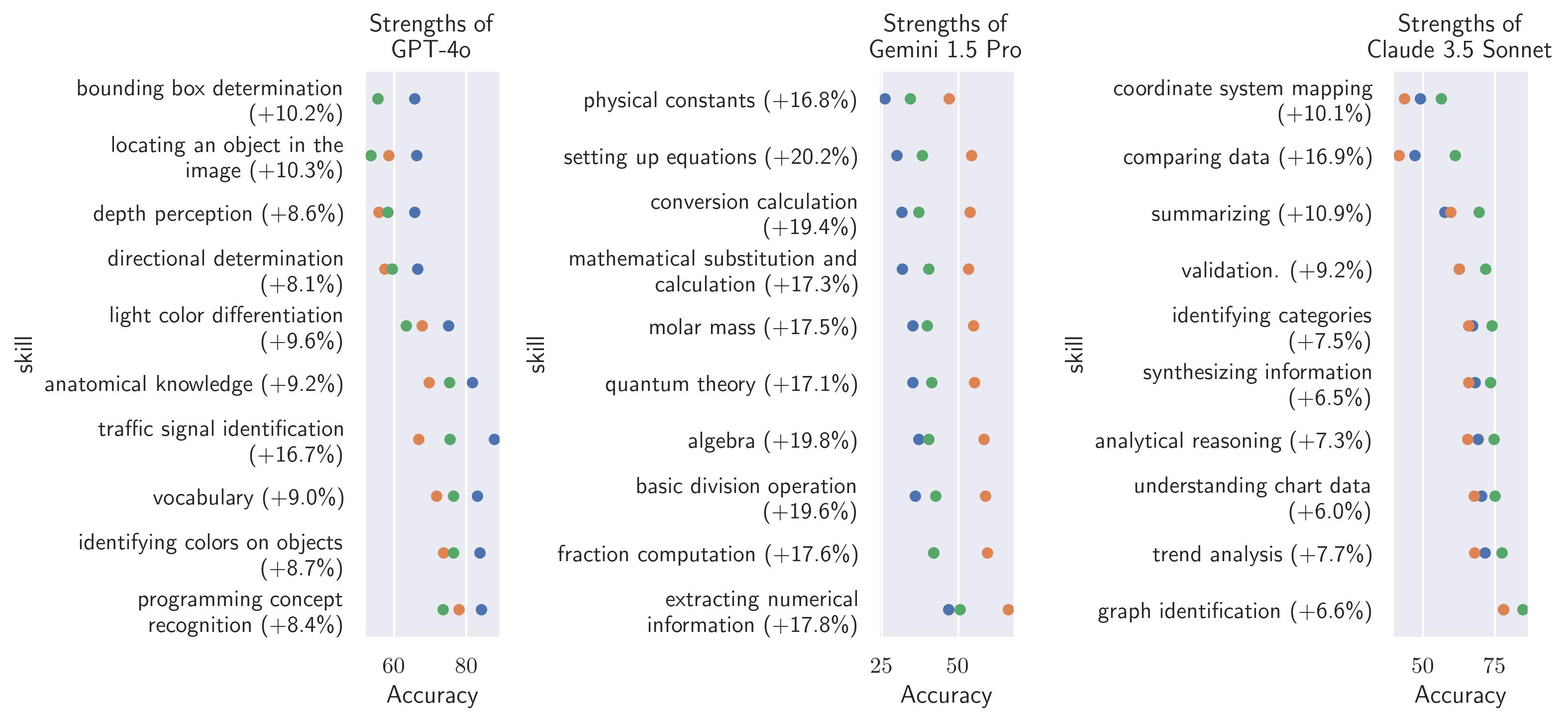}
    \includegraphics[width=0.5\linewidth, trim=0 10 0 10, clip]{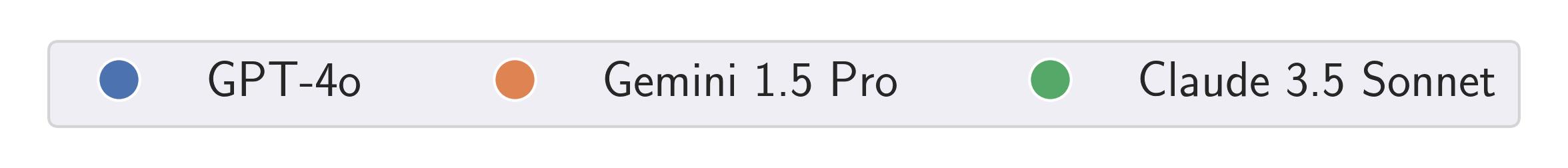}
    \includegraphics[width=1\linewidth]{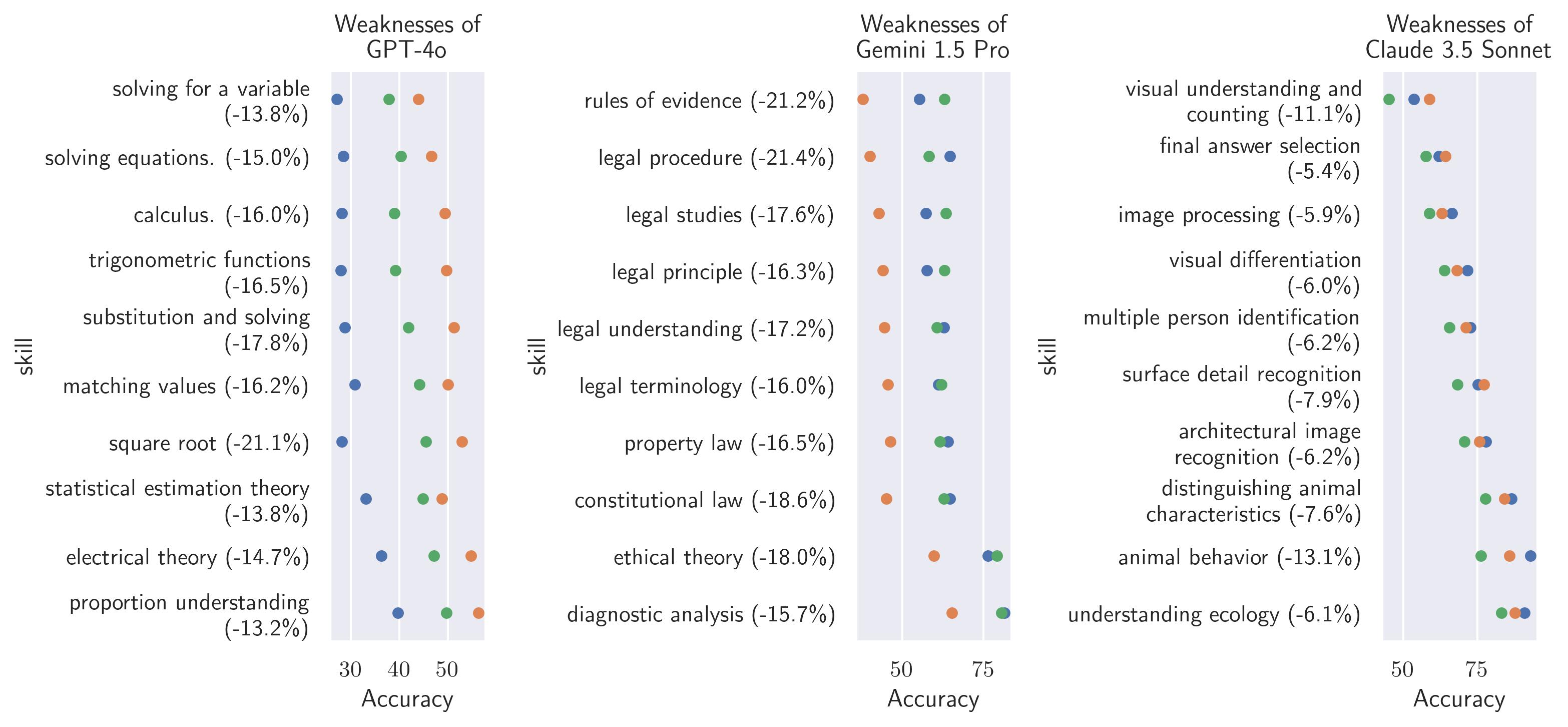}
    \caption{Unique strengths (\textbf{top}) and weaknesses (\textbf{bottom}) of GPT-4o, \GeminiPro, and \ClaudeSonnet, relative to one another. For each model, we present skills where the model's slice accuracy is highest / lowest (respectively) relative to the average of the other two model accuracies.}
    \label{fig:strengths-and-weaknesses}
\end{figure}


\textbf{Interpreting Trade-offs.} We now directly compare GPT-4o, \GeminiPro, and \ClaudeSonnet on all skill-slices with at least $100$ unique evaluation instances. \emph{While overall accuracies for the three models across our evaluation corpus falls within a half percent of one another, we observe differences as high as $20\%$ for certain skills}, as shown
in figure \ref{fig:strengths-and-weaknesses}. Further, some patterns emerge: \GeminiPro is distinctly strong for math and science, while GPT-4o's relative strengths pertain to visual skills like color differentiation and object localization, as well as the real-world skill of \emph{traffic signal identification}, where GPT-4o is on average $16.7\%$ more accurate. Interestingly, \GeminiPro's greatest relative weaknesses nearly all concern law, which we found above to be the key skills that GPT-4o and \ClaudeSonnet improved upon compared to their respective predecessors. 

\begin{tcolorbox}[width=\textwidth, boxsep=0.01mm]    
   Finding 2: \textbf{Models have unique strengths and weaknesses}:
   e.g., GPT-4o excels in visual skills, but lags behind on various math skills. Conversely, \GeminiPro outperforms both other models on math and science skills, while falling far behind for legal skills (figure \ref{fig:strengths-and-weaknesses}).
\end{tcolorbox} 

Unfortunately, since we have limited visibility into the training data and processes for these private models, we cannot offer explanations for the strengths and weaknesses we observe. Nonetheless, this illuminates an advantage of our method: by annotating evaluation instances, \textbf{skill-slice analysis enables fine-grain insight on model capabilities, even when only granted black-box access}.



\subsection{Generalization of Skill-Slice Insights}

Our skill-slice analysis operates under the premise that the insights drawn from inspecting sufficiently-large slices from a wide range of sources can \emph{generalize} to new instances. If the insights generalize well, then understanding skill trade-offs between models can enable employing the models in a more calculated manner. That is, if we have knowledge of the skills relevant to a new instance, as well as each model's skill-wise strengths, we can improve accuracy by \emph{routing} that instance to the model whose strengths are most aligned with those skills. 

\begin{figure}   
    \centering
    \begin{minipage}{0.49\linewidth}
        \centering
        \includegraphics[width=\linewidth]{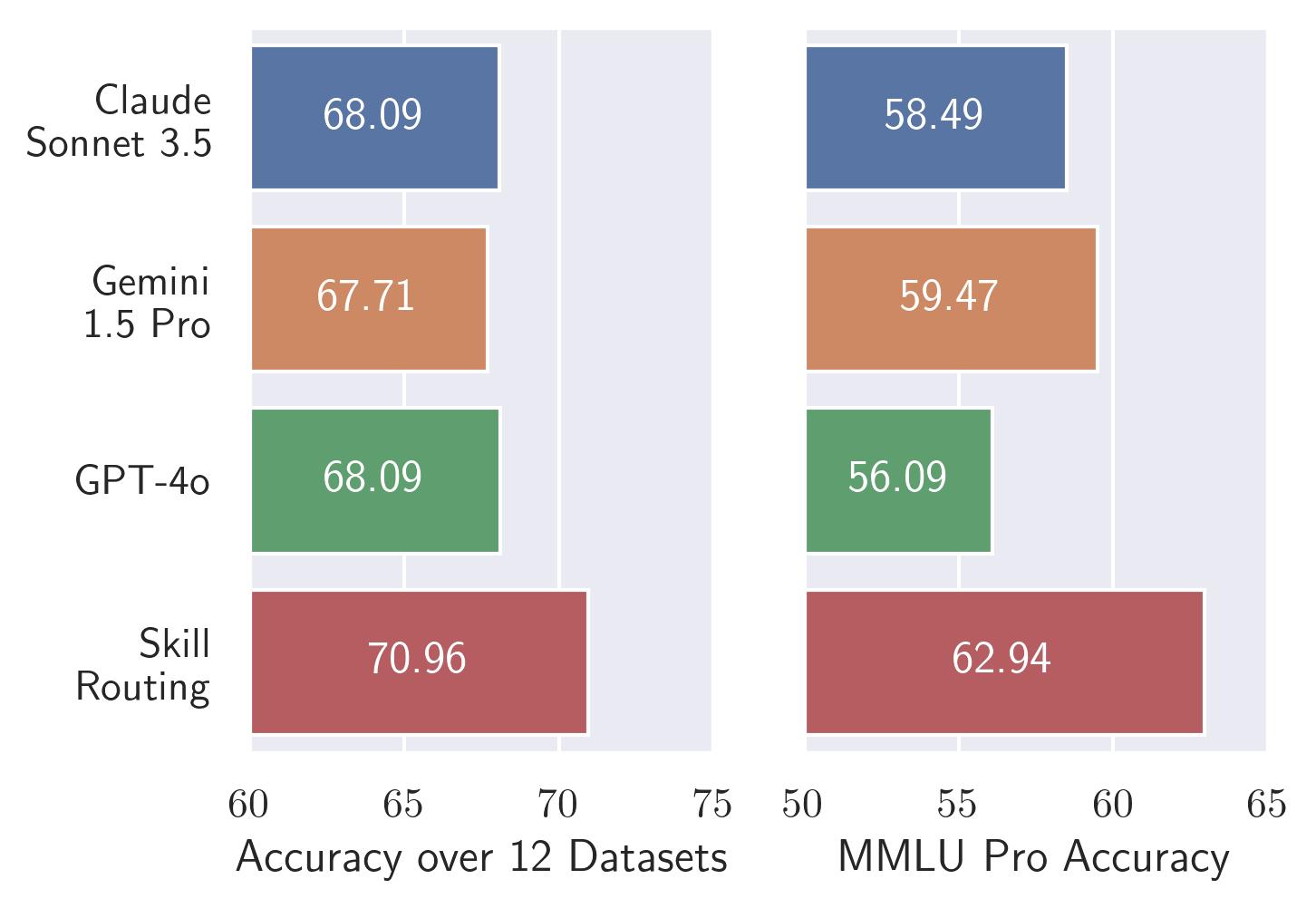}
    \end{minipage}
    \begin{minipage}{0.49\linewidth}
        \centering
        \includegraphics[width=\linewidth]{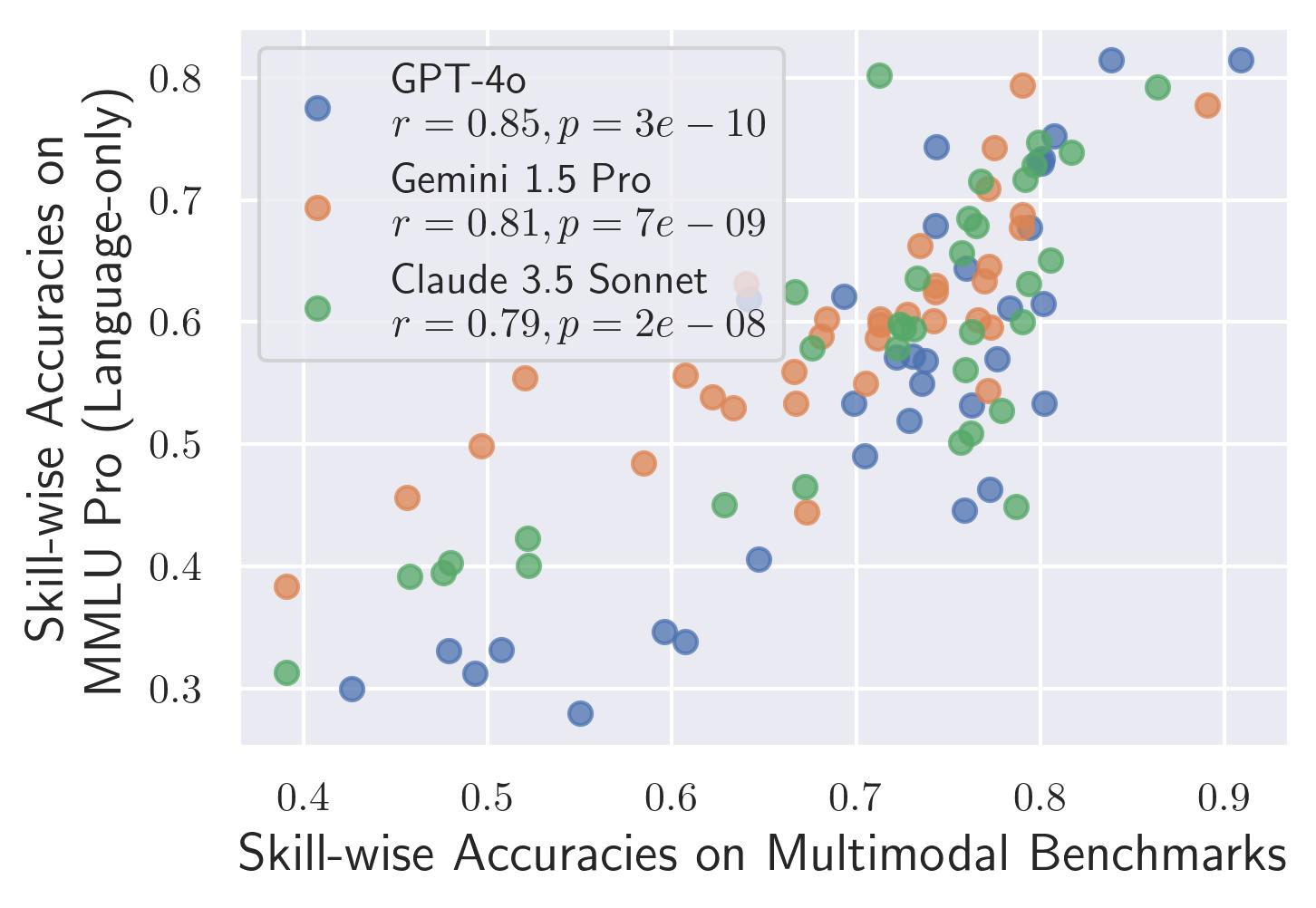}
    \end{minipage}
    \caption{(\textbf{left}) A simple routing scheme that assigns each instance to the model with best accuracies for the skills relevant to that instance can lead to noticeable accuracy gains. 
    (\textbf{right}) Accuracies over skill-slices can generalize to unseen data. In fact, skill-wise accuracies computed over multimodal benchmarks correlate strongly with accuracies obtained over a language-only benchmark.
    }
    \label{fig:skill-generalization}
\end{figure}

\looseness=-1
To test this, we route each instance in our corpus to one of GPT-4o, \GeminiPro, and \ClaudeSonnet, based on the skill annotations for that instance and the skill-wise accuracies per model computed over the remaining corpus (i.e. without the test instance). 
To obtain a single score per instance per model, we take a weighted average of skill-wise accuracies, where the weight for each skill is the inverse of its slice size (so to upweight finer-grained, more specific skills).
As shown in the left panel of figure \ref{fig:skill-generalization}, routing increases accuracy by up to $3.2\%$ compared to each of the frontier models alone on 12 datasets combined, including improvements of $3.5$ to $6.8\%$ for MMLU Pro \citep{wang2024mmluprorobustchallengingmultitask}. 

We highlight the MMLU Pro results because the vast majority of the instances in the reference corpus (over which skill-wise accuracies are computed) are from multimodal benchmarks, while MMLU Pro is language-only, making it a good candidate to showcase the generalization of our approach. To study this deeper, we now partition each skill-slice based on if the instance comes from MMLU Pro or one of our multimodal benchmarks. Then, for each model we obtain two paired sets of slice-wise accuracies: one accuracy score per skill-slice, per partition. As shown in the right panel of figure \ref{fig:skill-generalization}, we observe strong correlations ($r\geq0.79$, $p<1e-7$) between these two sets of scores for all three models. In addition to showing generalization, this result further validates our idea that skills pertain to a deeper property of a given instance (namely, what must be done to \emph{solve} it) than surface level attributes, like its modality. 


\begin{tcolorbox}[width=\textwidth, boxsep=0.01mm]    
   Finding 3: \textbf{Skill-slice accuracies enable instance-wise model selection}. Routing each instance to the model strongest on the relevant skills results in a $3\%$ accuracy gain, indicating that skill-slice analyses offer generalizable insight (see figure \ref{fig:skill-generalization}). 
\end{tcolorbox} 

We note that this proof of concept utilizes accuracies computed over slices defined by a single skill, not accounting for the added difficulty created when certain skills are used together. Nonetheless, our framework opens the door to studying these second order effects, e.g., by surfacing error-inducing slices defined by a \emph{combination} of skills, as done by \cite{rezaei2024prime} for image classifiers.

\section{Corroborating Slice Accuracies with Probing Questions}
\label{sec:probing}

\looseness=-1
While skill-slices allow for approximating a model's proficiency at a skill by averaging accuracy over any instance where the skill is relevant, our rationale parsing  framework enables a second independent analysis to more directly probe a \emph{single} skill, without the effect of co-occurring skills. 

\begin{figure}
    \centering
        \begin{minipage}{0.4\linewidth}
        \includegraphics[width=\linewidth]{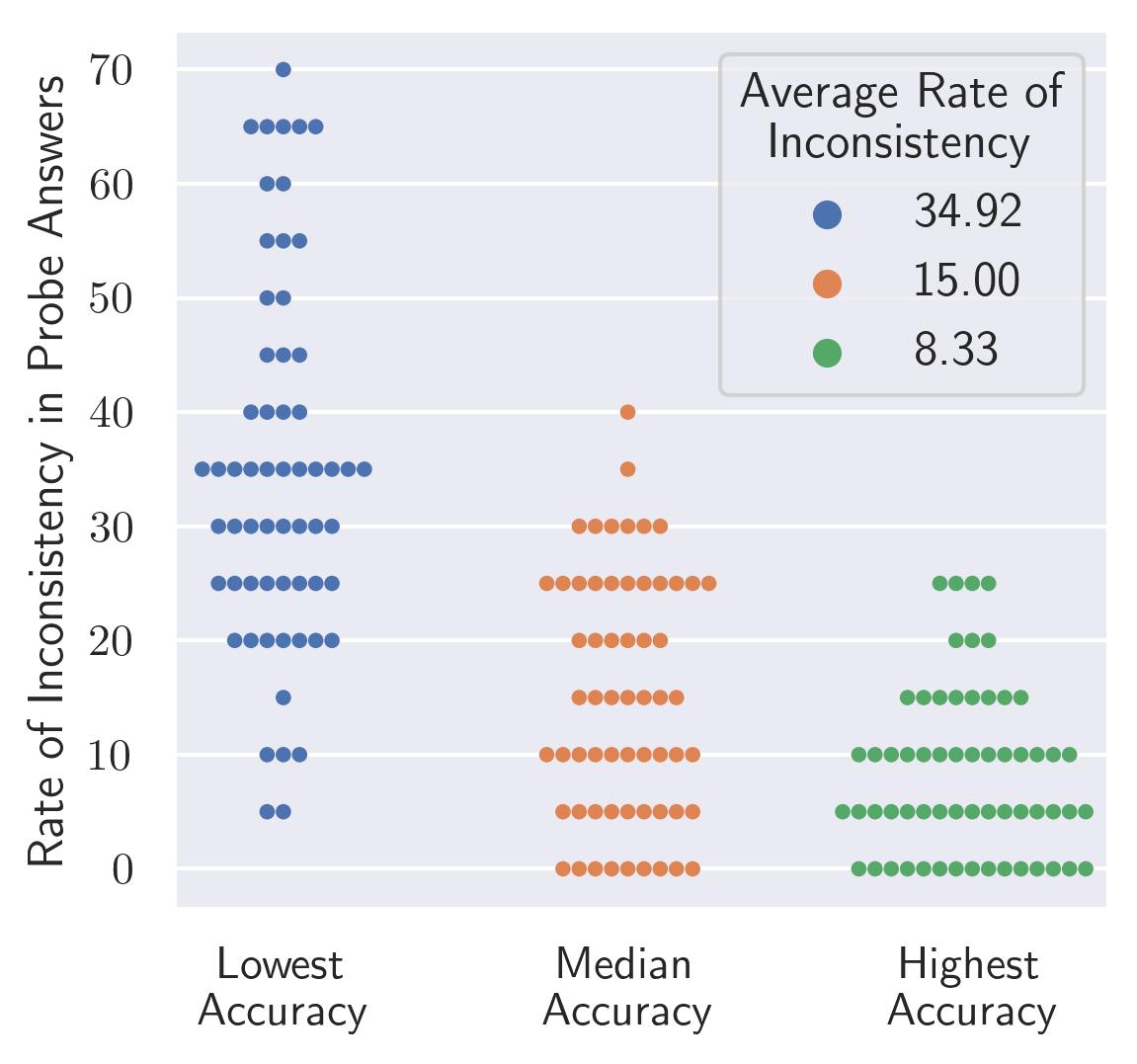}
        \end{minipage}
        \begin{minipage}{0.58\linewidth}
        \includegraphics[width=\linewidth]{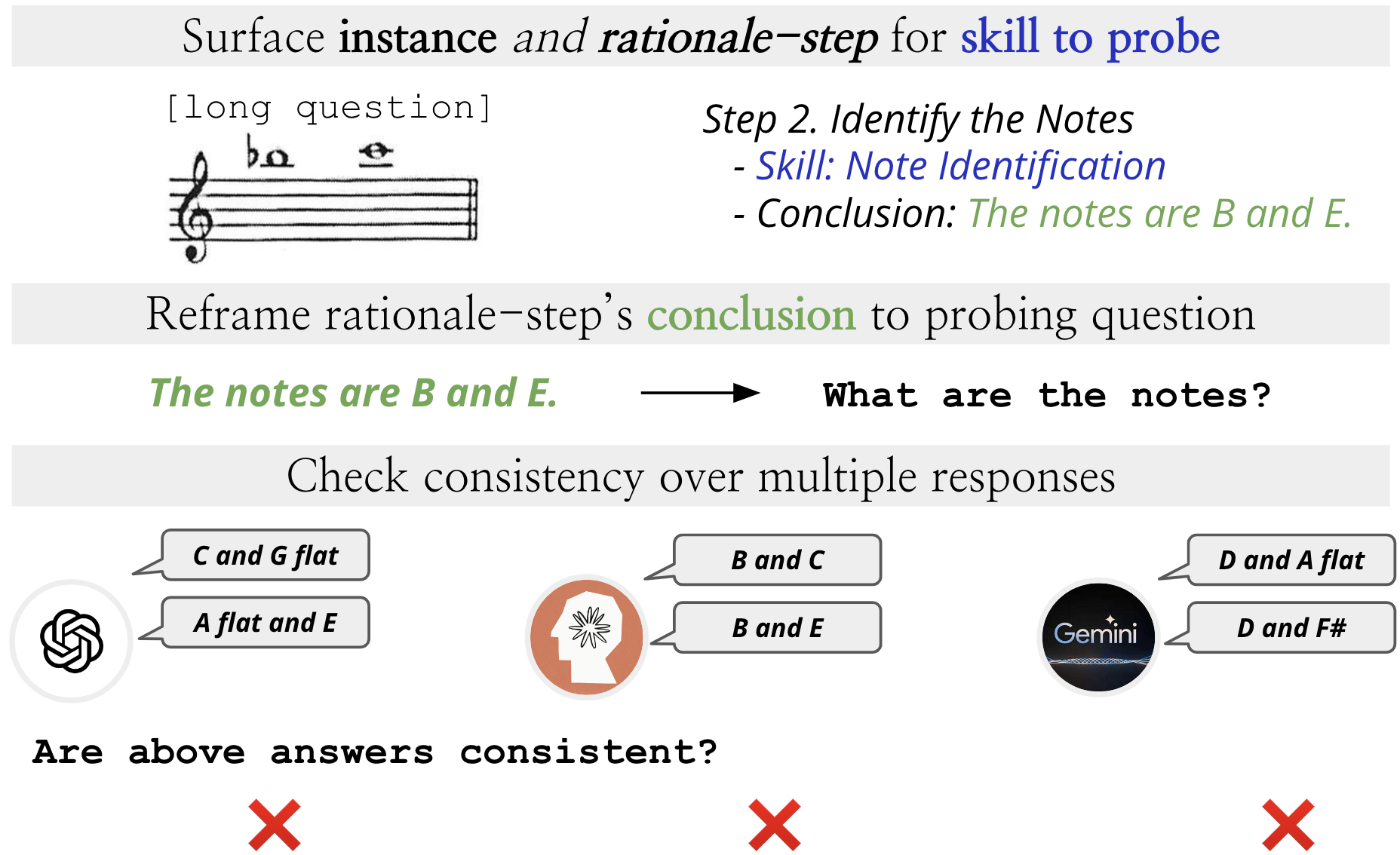}
        \end{minipage}
        \caption{(\textbf{left}) Skills who's slices yield lowest model accuracy have the highest rates of inconsistency when probed. Each point corresponds to one model's inconsistency rate for a single skill. (\textbf{right}) Probe questions are generated by reframing rationale steps associated with a skill of interest.}
    \label{fig:probe-consistency}
\end{figure}

\looseness=-1
Namely, rationale parsing \emph{localizes} each skill to a specific step of the generated solution, as well as the resulting claim, which we instruct the rationale-generating model to include in its response. For example, as shown on the right in figure \ref{fig:probe-consistency}, the resultant claim to a step where the skill ``[musical] note identification" is applied is ``The notes are B and E". A claim can then be reframed as a question (e.g. ``What are the notes?") probing a single skill, unlike the original question that often requires numerous skills and steps, any one of which could cause an incorrect answer. We use the rate of \emph{inconsistency}, like in \cite{wang2023selfconsistency}, over multiple responses to each probing question as a measure of proficiency at the probed skill: if a model contradicts itself, it must have been wrong at least once.

\looseness=-1
We now probe three sets of $20$ skills: those with lowest, median, and highest accuracy based on the analysis in section \ref{sec:findings}, to sample skills with a variety of slice accuracies. For each skill, we (i) generate $20$ probing questions, (ii) obtain $5$ responses per probe per model, and (iii) evaluate consistency of the responses with GPT-4o\footnote{While we use GPT-4o, forming probing questions and evaluating consistency are likely simple enough to be done reliably by open-source LLMs (as both tasks are text-only). See appendix \ref{app-sec:probing-prompts} for all prompts used.}. 
As shown in figure \ref{fig:probe-consistency}, models contradict themselves at a significantly higher rate for low accuracy skills, and \textbf{inconsistency rises as slice accuracy falls}, corroborating the skill-slice analysis. Overall, inconsistency correlates well with slice accuracy ($r=-0.675$; see Appendix \ref{app-sec:probing-corrs}). Further, we find skills where all models contradict themselves more often than not, such as ``[musical] note identification'', ``eye direction analysis'', and ``enumerating objects'', the last of which is a well-documented limitation of VLMs \citep{yuksekgonul2023when, Paiss_2023_ICCV}.

\begin{tcolorbox}[width=\textwidth,boxsep=0.01mm] 
   Finding 4: \textbf{Rationale-parsing enables targeting skills in isolation.} This independent analysis yields corroborating assessments of skill proficiency as skill-slice accuracies (figure \ref{fig:probe-consistency}). 
\end{tcolorbox} 

\looseness=-1
In addition to independently verifying the insights from slice accuracies, probe inconsistency offers \emph{complementary} signal: since some evaluation instances engage multiple skills, low accuracy over one skill-slice could be due to a deficiency on another highly co-occurring skill. However, if a model is proficient in a skill, it will always answer consistently (correctly) to probing questions. Thus, consistency on probing questions can be used to \emph{refine} insights from skill-slice analyses, as a skill with low slice accuracy and high probe inconsistency is almost surely deficient. We discuss automatic diagnosis of skill deficiencies further in Appendix \ref{app-sec:probing-e2e}, leaving it as a future application of our work.


\looseness=-1

\section{Retrieving Evaluation instances for Custom Query Skills}
\label{sec:retrieval}

\looseness=-1
Finally, towards (a) direct comparison of our work to prior methods and (b) more flexible use of skill annotations, we introduce the task of \emph{skill-based retrieval}: given an open-vocabulary query skill, can we retrieve relevant evaluation instances, so to build a custom skill-slice for specialized evaluation? This task boils down to defining an efficient metric that assigns a similarity score between a text query and an evaluation instance. Here, we focus on multimodal evaluation instances\footnote{A multimodal instance consists of one \textbf{image} along with a \textbf{text} question}. 

\looseness=-1
We consider \textbf{baselines} that leverage \textbf{a.} embeddings of the instance (either of just the image using CLIP \cite{clip}, just the text question, or the average of both image and text), \textbf{b.} attributes of the input, either inferred by GPT-4o\footnote{We obtain surface-level `tags': the type of image, subject of the question, etc. See App. \ref{app-sec:retrieval}.} or provided as ground-truths by benchmark creators. In this latter case, we embed text attributes, along with the query skill, using a text encoder, with which we compute similarity of an attribute to the query skill as the cosine similarity of their embeddings. To obtain a single similarity score for the GPT-4o attributes baseline where multiple attributes exist per instance, we average the top 3 highest similarities\footnote{As in \cite{embracing-diversity}, we find top-3 averaging to be superior to top-1 or full averaging.}. Our method is identical to the attribute-based approach, except that we leverage annotated \emph{skills} instead of attributes. The baselines encapsulate how prior methods propose to group inputs to go beyond overall accuracy \citep{eyuboglu2022domino, rezaei2024prime}, enabling direct comparison to our method. 

\begin{wrapfigure}{l}{0.5\linewidth}
    \vspace{-0.3cm}
    \centering
    \includegraphics[width=\linewidth]{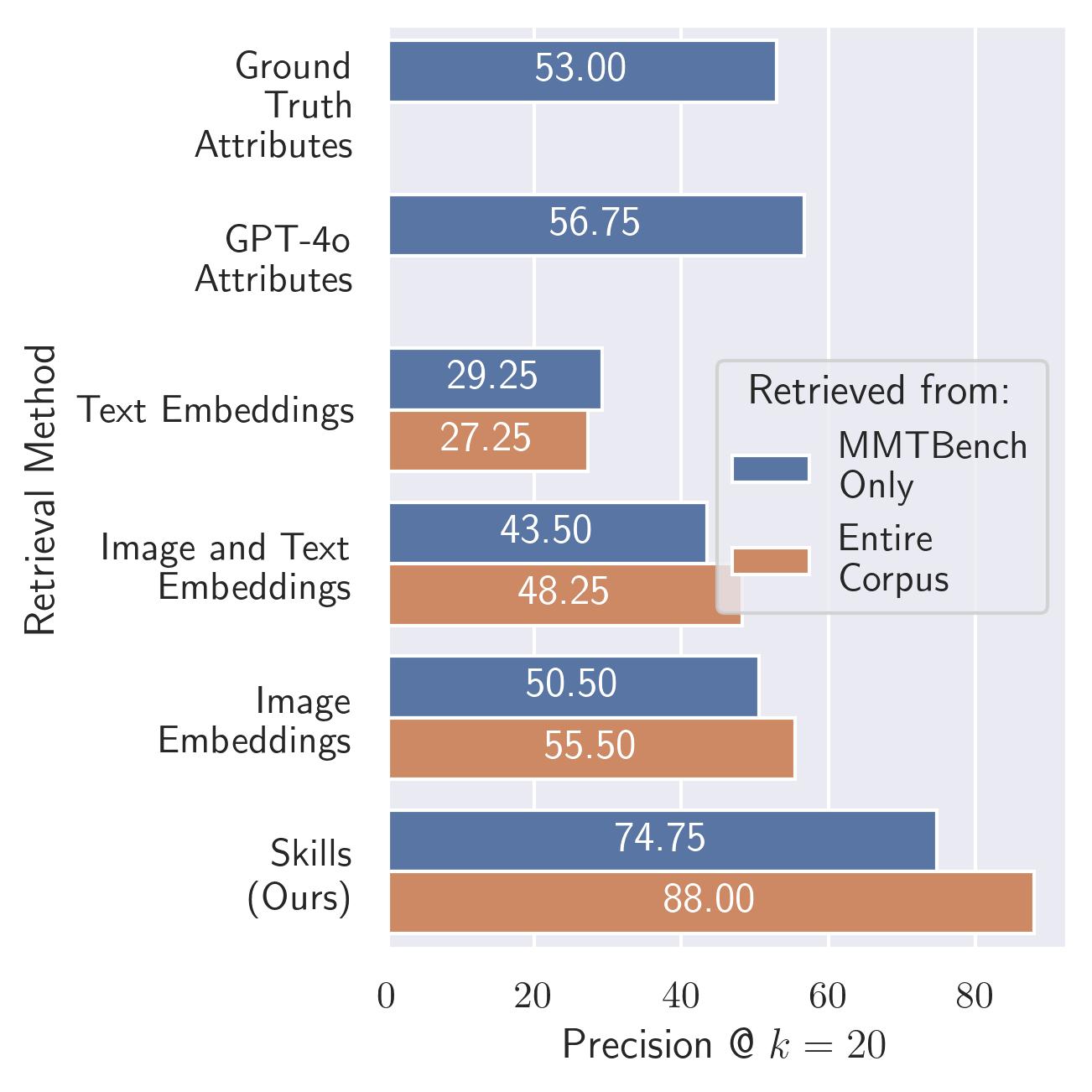}
    \caption{Retrieving instances based on a query skill is far more effective when utilizing skill annotations than using direct representations of each instance, as done in prior work.}
    \label{fig:skill-retrieval}
\end{wrapfigure}


\textbf{Setup:} We take 20 skills from one benchmark (MMBench), and retrieve $20$ instances per skill per method from two sources: \textbf{a.} a single other benchmark: MMTBench, selected because it has the highest number of ground-truth categories (162) out of our corpus, \textbf{b.} all benchmarks aside from MMBench, though we exclude ground-truth and GPT-4o inferred attributes baselines in the second case, for availability and cost reasons respectively. \textbf{Metric:} We evaluate each retrieved instance by presenting it along with the query skill to GPT-4o and prompting it to decide if the query skill is relevant or not; `precision @ $k=20$' is the fraction of the $20$ retrieved instances deemed relevant. 

As shown in figure \ref{fig:skill-retrieval}, retrieval using skill annotations far exceeds the precision for any other baseline in both settings, with improvements of $18$-$45.5\%$ and $50.75$-$32.5\%$ respectively. 
The attribute baselines are stronger than the embedding-based approaches, though obtaining attributes is also more expensive. 
The embedding-based baselines struggle, reflecting that frozen-embedding models may prioritize encoding surface-level information over skills, as skills may only become more apparent when inspecting the \emph{solution} to the instance. We also find that searching over a broader corpus benefits the skill-based approach far more than either embedding-based approach. In summary, while allowing for direct searching over our corpus, these skill-based retrieval experiments underscore how our skill annotations offer novel signal, enabling the grouping of instances in ways that can be complementary to prior methods. 

\begin{tcolorbox}[width=\textwidth,boxsep=0.01mm] 
   Finding 5: \textbf{Skill annotations open the door to custom evaluation sets via skill-based retrieval.} Similarity of attributes or direct embeddings for two instances does not reflect the similarity of the underlying skills tested by each instance, leading to poor skill-based retrieval performance (figure \ref{fig:skill-retrieval}). 
\end{tcolorbox}

\section{Review of Literature}
\vspace{-2mm}
Prior efforts to gain \textbf{fine-grain insights} often involve intensive manual efforts to construct multi-task benchmarks \citep{liang2023holistic, superglue, srivastava2023beyond, altahan2024unibenchvisualreasoningrequires, eureka} or datasets with annotations beyond ground-truth labels \citep{white2024livebenchchallengingcontaminationfreellm, fu2023mme, yu2024mmvet, lu2024mathvista}. On top of quantifying specific abilities, extra annotations have illuminated biases \citep{gendershades, worldbench} and robustness issues \citep{imagenetx, wilds2021}. Automated alternatives include synthesizing benchmarks with known attributes \citep{zhang2024task, bordes2023pug} or automatically uncovering sub-groups within existing data \citep{luo2024llmdatasetanalystsubpopulation, qualeval}.
 The term \emph{slice discovery} (or otherwise referred to as data \emph{cohort} or \emph{subgroup}) was coined \citep{chung2019slice,eyuboglu2022domino} to describe these methods which enable comparisons \citep{changelists} and error analysis \citep{rezaei2024prime,vlslice,nushi2018towards,singla2021understanding,dunlap2024describing}. 
While these works focus on slices formed by surface-level attributes for visual recognition tasks, we curate \emph{skill-slices}, which pertain to the underlying ability each slice requires from a model, and as such extend to more tasks and modalities.


\looseness=-1
\textbf{Skills} are akin to the `ability' tags present in some modern benchmarks \citep{liu2024mmbench, mmtbench}, though they can be finer (or coarser) grained \citep{learning_hierarchies} than what is typically annotated. The importance of acquisition of skills has been studied in human cognition \citep{robert_m__gagne_1962, koedinger} as well as language models \citep{arora2023theoryemergencecomplexskills}. \cite{yu2024skillmix} utilizes skills as input to an LLM to generate challenging evaluation instances requiring a composition of the provided skills. Recent \citep{qualeval} and concurrent \citep{metacog} works also explore skill inference, albeit in more narrow domains, toward improved prompting or finetuning. This important shift of methods from attributes to skills is indeed motivated by the increasingly general-purpose nature of state-of-the art models. In this work, we scale up such analysis and insights by expanding it to 12 benchmarks and by introducing \emph{rationale parsing} as a means to infer more skills.


\textbf{Rationales} have been studied extensively in the context of prompting \citep{wei2022chain, kojima, yao2023tree}. Other works show the value of rationales as richer training signal \citep{hsieh2023distilling, orca2, zelikman2022star, krishna2023post} or model explanations \citep{xai_rationales, Hu_Yu_2024, huang2023largelanguagemodelsexplain}, though some question their faithfulness to underlying model processes \citep{madsen-etal-2024-self, fayyaz2024evaluatinghumanalignmentmodel}. We utilize rationales not to interpret a model directly, but instead to improve the relevance and diversity of the annotated skills \citep{Singh2024RethinkingII}, which become apparent in the solution steps present in rationales.

\section{Discussion and Future Work}
\vspace{-2mm}
It is often said, ``We can only improve what we can measure." As models grow in complexity and capability, traditional evaluation metrics like aggregate accuracy over entire datasets become insufficient. Simply averaging accuracy across datasets hides important insights needed to understand and enhance both the datasets and the models themselves.

To address this gap, our work introduced a skill-slice analysis approach for model evaluation. By examining model-generated rationales, we developed an automated method to extract the underlying skills required to solve individual evaluation instances. This shift from focusing on overall accuracy to conducting a more detailed analysis provides a more granular picture of a model's strengths and weaknesses. Furthermore, our reliance on a strong model to help humans understand evaluation data may prove to be increasingly pertinent as models get so strong that only a small handful of people will be qualified to provide manual annotations on evaluation instances. In this sense, our work relates to scalable oversight \citep{Bowman2022MeasuringPO}, providing a new avenue to evaluation leveraging human-AI collaboration to understand models that will encapsulate advanced skills and knowledge.

\looseness=-1
Looking ahead, our approach can assist in creating more focused, skill-based evaluation datasets, by accelerating the process of discovering skills where models are collectively deficient or exhibit disparate performance. Our skill-based retrieval method can play a pivotal role in generating valuable training data for model development. By concentrating on specific skills, one can create more effective training sets that address identified weaknesses, thereby accelerating the advancement of AI models. This aligns with the principle that meaningful progress is achievable when we have clear insights, reinforcing the notion that we can only improve what we can measure.

\section{Acknowledgements}
We sincerely thank Siddarth Joshi, Alessandro Stolfo, and Natasha Butt for invaluable conversations and feedback throughout this project.

\newpage

\bibliography{iclr2025_conference}
\bibliographystyle{iclr2025_conference}

\appendix
\newpage
\section{Limitations}
\label{app-sec:limitations}

A (somewhat surprising) key hypothesis our work hinges on is that models can articulate relevant skills, even for evaluation instances that they fail to answer correctly. While we empirically validate this claim for (a subset of) the skill annotations we present, it is possible that on harder benchmarks, the listed skills may no longer be accurate. Thus, we recommend always employing at least our post-hoc verification when collecting skill annotations on a new set. 

Perhaps what is more likely is that while listed skills are relevant, they may be incomplete. Note that we do not present evidence for completeness in the main text. High overlap between two skill annotators is a promising sign (see Appendix \ref{app-sec:validation}), though the fact that overlap is not perfect suggests some skills may be missed. Moreover, most skills can always be broken down into smaller skills, making completeness perhaps an infeasibile and extraneous objective.

Lastly, generating rationales is more expensive than directly prompting a strong model to list skills, or simply using a encoder to embed an entire instance at once. We show rationales have some advantages compared to each of these approaches in Appendix \ref{app-sec:prompt-ablation} and Section \ref{sec:retrieval} respectively, though the added cost of our method is worth mention.

\section{Composition of Skill-Slices in our Study}
\label{app-sec:skill-set}

Here, we enumerate the datasets we compute skill annotations for, as well as provide more details on the resultant slices we form. All annotations and slices, which we call the \emph{Skill-Index}, will be released to the broader community to enable skill-level analyses on models of their choice. We will also release code to easily add to the \emph{Skill-Index}.

\textbf{Datasets}. We include datasets from 12 benchmarks in our study, consisting of 11 multimodal (image and text) datasets and 1 language-only dataset. These datasets are:
\begin{itemize}
    \item MMLU Pro, a language-only benchmark by \cite{wang2024mmluprorobustchallengingmultitask}, intended to be a harder version of MMLU \citep{hendryckstest2021}
    \item MMMU (val), a multimodal benchmark with college level questions from many academic subjects intended to test `expert' AI, by \cite{yue2023mmmu}
    \item MathVista, a mathematics visual understanding benchmark by \cite{lu2024mathvista}
    \item MMC, a chart understanding multimodal benchmark by \cite{liu-etal-2024-mmc}
    \item MMVP, a benchmark specifically focusing on failure modes of VLMs, by \cite{liang2024eyes}
    \item Many general multimodal benchmarks testing numerous abilities:
    \begin{itemize}
        \item MMBench, by \cite{liu2024mmbench}
        \item MMTBench, by \cite{mmtbench}
        \item MME, by \cite{fu2023mme}
        \item MMVet, by \cite{yu2024mmvet}
        \item SEEDBench, by \cite{li2023seedbench}
        \item Realworld-QA, a benchmark that claims to test many real-world visual understanding questions, by \cite{realworldqa}
        \item VibeEval, by \cite{padlewski2024vibeeval} (also referred to as reka\_vibe, as it was produced by the company Reka).
    \end{itemize}
\end{itemize}
The two largest datasets in our benchmark suite are SEEDBench and MMLU Pro, which have $14$k and $12$k instances respectively. 

We primarily sought out multimodal benchmarks, as many detailed benchmarks exist for language-only tasks. We aimed to select popular benchmarks by prioritizing benchmarks that were featured in recent reports for model releases or those that had a large number of recent downloads on huggingface. We stress that benchmarks can easily be added to our corpus -- our experiments suggest our prompt is effective in diverse settings. 

\textbf{Forming skill-slices}. To consolidate the $128$k unique total skills we extract, we perform a tight clustering over embeddings of the skill names. We use the \texttt{SFR-Embedding-2\_R} model \citep{SFR-embedding-2}, as it was the leader on the huggingface MTEB leaderboard at the time \citep{muennighoff2022mteb}. We utilize the fast clustering algorithm from \texttt{sentence-transformers} \citep{reimers-2019-sentence-bert} with a minimum similarity threshold of $0.95$, intended to de-duplicate skills. One could use a lower threshold, though in that case, naming a resultant skill cluster would be more challenging, as the skills within a cluster would be more disparate. At our threshold, we qualitatively observe nearly all skills within a cluster have very similar names, so we simply name the skill cluster with the skill in the cluster who's embedding is closest to the centroid. To be sure we have completely de-duplicated, we run the clustering algorithm repeatedly (about 2-3 times) until all but at most $0.5\%$ of cluster names have a similarity below $0.95$ to any other cluster name. 

Throughout our paper, we use skill-slices with at least $100$ instances, though this number is arbitrary. Picking a lower threshold increases slice count (often including finer-grained slices), but accuracy over theses slices may be less reliable. 

\section{Details on Skill Extraction and Validation}
\label{app-sec:prompt}

We present the prompt used to generate rationales below. It features detailed instructions and an in-context example. We generate this prompt by iteratively asking GPT-4o to refine the prompt based on specific desiderata, along with manual tweaks along the way. The in-context example is also generated by GPT-4o; it was produced using an earlier version of the prompt, and then modified slightly to correct some errors. Note that by GPT-4o, we always mean \GPTFourO.

\begin{center}
\resizebox{0.75\textwidth}{!}{
\centering
\begin{tcolorbox}[colback=blue!1!white, colframe=blue!75!black, title=Prompt to generate rationales (part 1)]
\textbf{System Prompt: Detailed Step-by-Step Response with Skills Labeled}

\textbf{Objective:} List the skills and evidence for each step to solve the given question. Do so by responding to each question in a detailed, step-by-step manner, where each step utilizes only a single skill. \textit{Clearly state the skill being used and the evidence applied at each step.}

\textbf{Guidelines:}
\begin{enumerate}
    \item \textbf{Single Skill Per Step:}
    \begin{itemize}
        \item Ensure each step involves only one skill. Skills can be categorized with multiple names, each getting more specific. For example:
        \begin{itemize}
            \item \textbf{Knowledge:}
            \begin{itemize}
                \item Music Theory, Scale Identification, Reciting F Major scale
                \item Biology, Photosynthesis Process, Calvin cycle
                \item History, World War II Events, Battle of Stalingrad
            \end{itemize}
            \item \textbf{Perception:}
            \begin{itemize}
                \item Visual Recognition, Musical Note Identification
                \item Chart understanding, information extraction, determining the length of a bar in a bar chart
                \item Visual understanding, Spatial reasoning, Understanding depths of objects
                \item Auditory Recognition, Sound Identification, Recognizing a child's voice
                \item Video Analysis, Episodic memory, Remembering where an object was placed
            \end{itemize}
            \item \textbf{Reasoning:}
            \begin{itemize}
                \item Logical Deduction, Process of Elimination
                \item Mathematical Calculation, Algebraic Manipulation, Resolving fractions
            \end{itemize}
        \end{itemize}
    \end{itemize}
    \item \textbf{State the Skill:} Clearly identify and state the skill being used in each step.
    \item \textbf{List all pieces of relevant evidence:}
    \begin{itemize}
        \item Specify the evidence used in each step. This could include:
        \begin{itemize}
            \item Information from the question (e.g., a specific phrase).
            \item Data or findings from previous steps. \textit{List these findings clearly if they are required context for the claim of the step!}
            \item Regions in an attached image, if applicable, \textit{clearly stating WHERE in the image the information is being extracted from.}
        \end{itemize}
    \end{itemize}
    \item \textbf{State the Conclusion of the Step.}
    \item \textbf{List your final answer after your step-by-step explanation in a new line formatted as 'ANSWER: \{\}'}
\end{enumerate}

Here is an example of the desired response structure given a question.
\end{tcolorbox}}
\end{center}

\begin{center}
\resizebox{0.75\textwidth}{!}{
\begin{tcolorbox}[colback=blue!1!white, colframe=blue!75!black, title=Prompt to generate rationales (part 2)]
\textbf{Example Question:} "Based on the following image, can you determine what the two sounds belong to? \textbackslash image 1\\ A. \#F major scale\\ B. c melodic minor scale\\ C. b harmonic minor scale\\ D. E melodic major scale"

\textbf{Example Response Structure:}
\begin{enumerate}
    \item \textbf{Step 1: Recognize the Type of Image}
    \begin{itemize}
        \item \textbf{Skill:} Perception: Visual Recognition, Image Classification, Sheet music recognition
        \item \textbf{Evidence:} The image has lines and symbols, which resemble sheet music.
        \item \textbf{Conclusion:} The image is of sheet music.
    \end{itemize}

    \item \textbf{Step 2: Identify the Clef}
    \begin{itemize}
        \item \textbf{Skill:} Perception: Visual Recognition, Symbol Identification, Treble clef recognition
        \item \textbf{Evidence:} There is a staff in the middle of the image, and on the left, there is a treble clef.
        \item \textbf{Conclusion:} The music shown is on the treble clef.
    \end{itemize}

    \item \textbf{Step 3: Identify Symbol Positions from the Image}
    \begin{itemize}
        \item \textbf{Skill:} Perception: Visual Recognition, Spatial reasoning, Locating musical symbols 
        \item \textbf{Evidence:} Two notes are present in the middle of the image: one with a sharp in the second space from the bottom, and one in the space above the top line.
        \item \textbf{Conclusion:} The notes are F\# and A.
    \end{itemize}

    \item \textbf{Step 4: Read Provided Options}
    \begin{itemize}
        \item \textbf{Skill:} Knowledge: Reading Comprehension, Understanding the question
        \item \textbf{Evidence:} The provided options are written after the question.
        \item \textbf{Conclusion:} The options are F\# major scale, c melodic minor scale, b harmonic minor scale, and E melodic major scale.
    \end{itemize}

    \item \textbf{Step 5: Eliminate Incorrect Options}
    \begin{itemize}
        \item \textbf{Skill:} Reasoning: Logical Deduction, Process of Elimination
        \item \textbf{Evidence:} 
        \begin{itemize}
            \item The notes shown are F\# and A, so they must be in the correct scale.
            \item F\# major includes F\# and A.
            \item c melodic minor does not include F\#.
            \item b harmonic minor does not include F\#.
            \item E melodic major is not a standard scale.
        \end{itemize}
        \item \textbf{Conclusion:} c melodic minor, b harmonic minor, and E melodic minor are incorrect.
    \end{itemize}

    \item \textbf{Step 6: Conclude the Correct Scale}
    \begin{itemize}
        \item \textbf{Skill:} Reasoning: Logical Deduction, Conclusion
        \item \textbf{Evidence:} F\# major scale is the correct answer, as it includes both F\# and A.
    \end{itemize}

    \textbf{ANSWER: A}
\end{enumerate}
\textbf{Final Note:} Ensure that the skills are described in detail: \textbf{LIST AT LEAST THREE NAMES FOR EACH SKILL.}
\end{tcolorbox}}
\end{center}

\begin{figure}
    \centering
    \includegraphics[width=\linewidth]{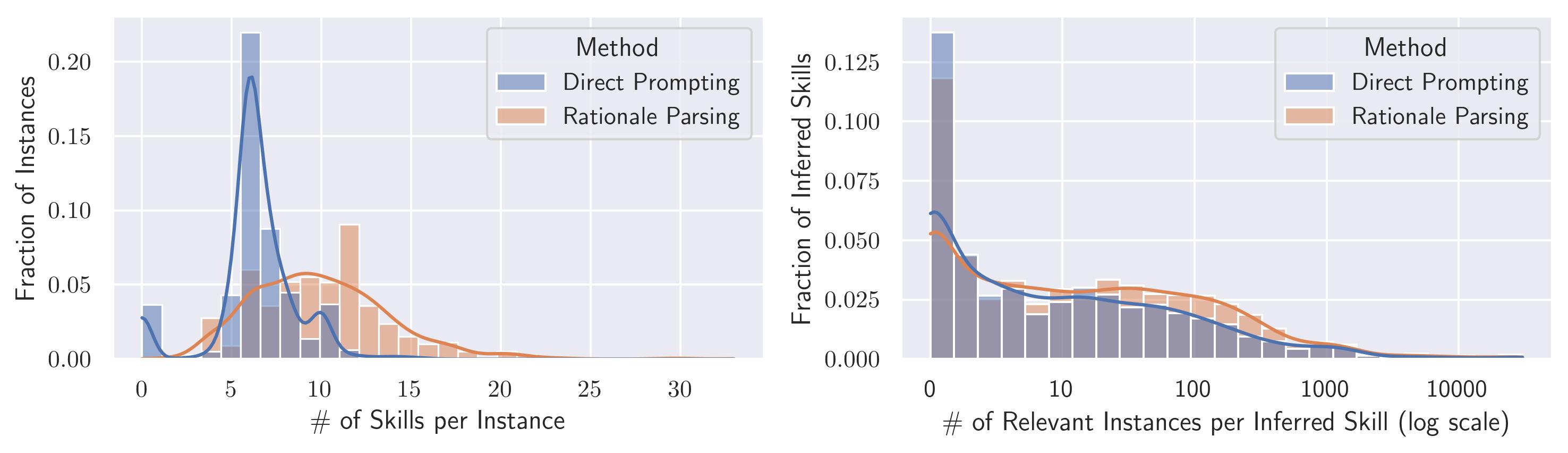}
    \caption{(\textbf{left}) Rationale parsing results in a significantly higher number of annotated skills per instance than using direct prompting. (\textbf{right}) Skills inferred with direct prompting are generally finer-grained, resulting in a higher fraction of very small slices (e.g. with $<10$ instances), and a lower fraction of all other slice sizes (see table \ref{tab:prompt_ablation} for exact numbers).}
    \label{fig:prompt-ablation}
\end{figure}

\subsection{Ablations on Rationale Parsing for Extracting Skills}
\label{app-sec:prompt-ablation}

\begin{table}[]
    \centering
\resizebox{\textwidth}{!}{
\begin{tabular}{lccccccc}
\toprule
$\#$ of Relevant Instances per Skill & $[0,10)$ & $[10,25)$ & $[25,50)$ & $[50,100)$ & $[100,250)$ & $[250,1000)$ & $\geq 1000$ \\
Skill Inference Method & & & & & & & \\
\midrule
Direct Prompting & $2494$ & $525$ & $341$ & $318$ & $288$ & $185$ & $123$ \\
Rationale Parsing & $2399$ & $560$ & $451$ & $409$ & $491$ & $290$ & $164$ \\
\bottomrule
\end{tabular}
}
    \caption{Comparison of granularity of skills inferred via direct prompting vs. rationale parsing. Direct prompting results in a higher number of skills that are very specific to the question, resulting in more skills with less than $10$ relevant instances than rationale parsing. Aside from this hyper-fine grain, rationale parsing results in more skills at every level of granularity, with largest improvements in medium grain slices. Details in section \ref{app-sec:prompt-ablation}.}
    \label{tab:prompt_ablation}
\end{table}

We now perform an ablation study to compare how skills inferred by  rationale parsing (our proposed method) differ from those obtained by directly prompting a strong model to list relevant skills. Specifically, the `direct prompting' alternate method consists of presenting an evaluation instance along with the prompt ``List skills that are relevant to the given question. Do not answer the question, only provide the list of skills. Be detailed, but only use short specific phrases throughout your response -- only use a few words per list item. I will provide you with an example first.'' As the prompt suggests, we additionally provide an in-context example, primarily to ensure standard response format to facilitate automatic parsing of the skill annotator model's outputs. We compare the annotated skills produced by GPT-4o over $1195$ instances from all $12$ benchmarks (roughly $100$ per benchmark) using direct prompting and rationale parsing. 

First, we observe that \textbf{rationale parsing results in a substantially higher number of unique skills annotated per instance than direct prompting}. The left panel of figure \ref{fig:prompt-ablation} shows the distributions of number of annotated skills per instance with the prompting methods. On average, we obtain $9.98$ skills per instance with rationale parsing, a $58\%$ relative improvement compared to the average of $6.32$ skills per instance with direct prompting. We note that for about $4\%$ of instances, direct prompting leads to $0$ inferred skills because the skill annotator directly answers the instance instead of listing the relevant skills. In the rationale parsing method, the skill annotator both answers the given instance and list skills simultaneously, alleviating issues with conflicting instructions. 

Next, we seek to compare the granularity of inferred skills from the two methods. Recall that there is no single `optimal' granularity: finer-grained skills are more descriptive, while coarse-grained skills result in larger skill-slices, which in turn improves the reliability of an average accuracy taken across such a slice. Thus, ideally, we can obtain skill annotations of various granularities. However, skills that are exceedingly fine-grained cannot be used, as skill-slices consisting of only a few instances lack the sample size needed to offer reliable measurements of model proficiency. 

To proxy a skill's granularity, we count the number of instances in our corpus annotated (via rationale-parsing) with a highly similar skill (i.e. similarity of at least $0.95$ between text embeddings of the skill names). Figure \ref{fig:prompt-ablation} (right) and table \ref{tab:prompt_ablation} show the distributions of the $\#$ of relevant instances per skill inferred via direct prompting and rationale parsing. We observe a larger fraction of skills inferred with direct prompting to be relevant for a very small number of instances, suggesting that \textbf{many of the skills inferred by direct prompting may be too fine-grained to result in sufficiently large skill-slices}. At other levels of granularity, rationale parsing consistently yields more skills for each grain. For example, $491$ skills inferred with rationale parsing have between $100$ and $250$ relevant instances, while only $288$ skills inferred with direct prompting have the same number of relevant instances --  a reduction of $41\%$. We  note that the improved diversity of grain of inferred skills using rationale parsing vs. direct prompting may in part be attributed to our instruction to list skills with multiple names of cascading granularity.


In summary, our proposed rationale parsing method results in more annotated skills per instance and greater diversity in granularity of annotated skills, when compared to direct prompting. Also, we qualitatively observe both methods to yield skills that are overwhelmingly relevant to the given instance -- we do not quantitatively test this facet, as we do not expect much difference in the relevance of skills annotated by either method. Another key advantage of rationale parsing is that it enables skill localization, which offers a second, complementary avenue to corroborate and refine our skill-slice findings (see sections \ref{sec:probing} and \ref{app-sec:probing-e2e}). 

We emphasize that it is certainly possible that other, perhaps better methods exist to extract skills. In this paper, we present one method and demonstrate the utilities it may hold. Many of these utilities can be extended and potentially expanded by improving our skill extraction method. Having demonstrated that skill inference is possible and useful, we leave further iteration on skill inference and its applications to future work. 


\subsection{Details on automatic validation}
\label{app-sec:validation}

\begin{figure}
    \centering
    \includegraphics[width=1\linewidth]{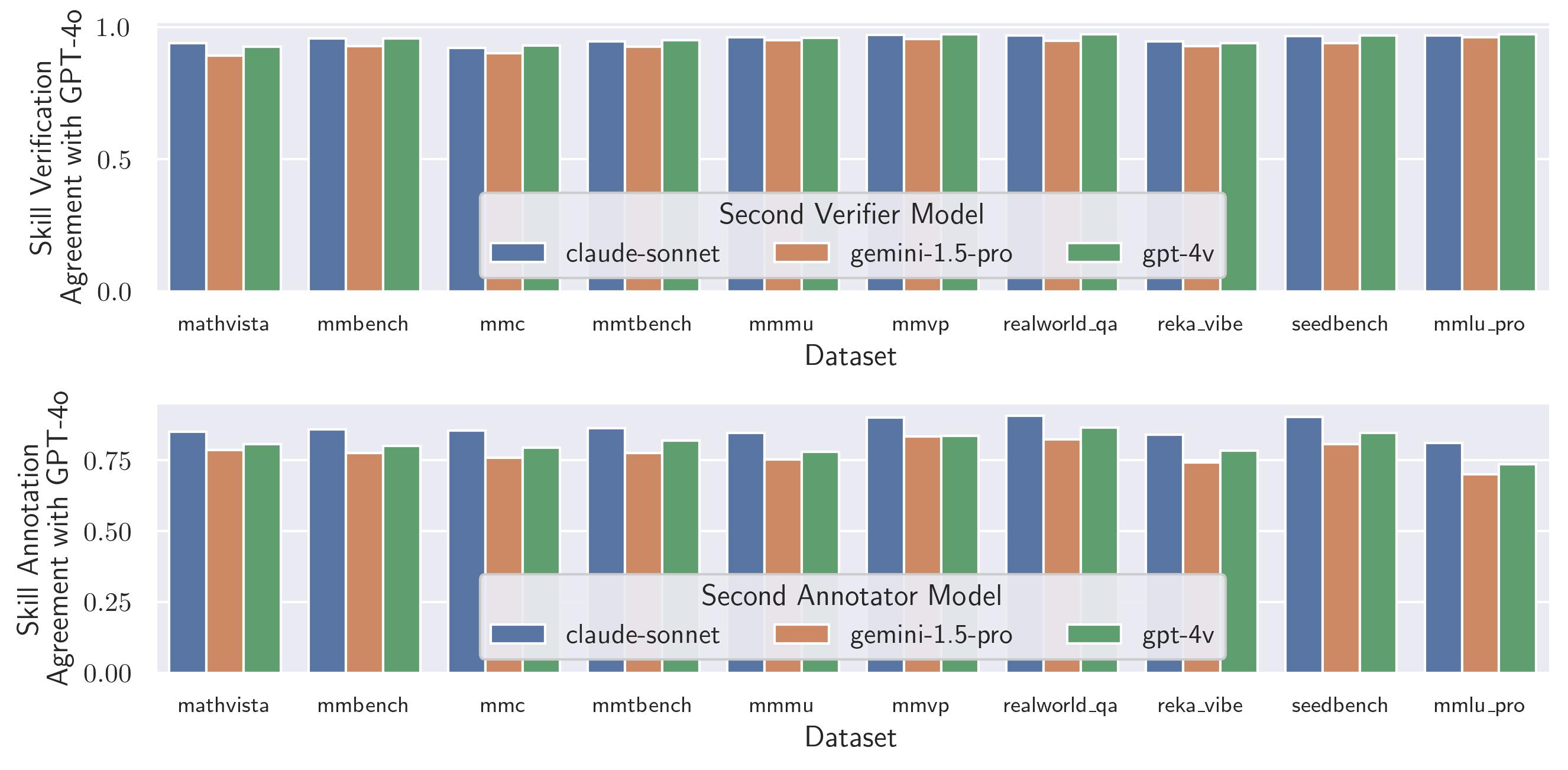}
    \caption{Diverse models agree when verifying the relevancy of an annotated skill (\textbf{top}) and annotating skills given an instance (\textbf{bottom}).}
    \label{fig:consistency}
\end{figure}

We now detail our second automatic validation approach: inter-(skill)annotator agreement. As the name suggests, we simply extract skills by feeding the same prompt to a new rationale-generating model. Namely, we try \ClaudeSonnet, \GeminiPro, and \GPTFourTurboApril. We compute agreement between two sets of skills listed by different models as the overall fraction of skills such that a match exists in the other set of skills. Here, we define two skills being matched if their text embeddings have a cosine similarity of at least $0.85$. Note that this is also the threshold we set for $\tau$ in sampling `negative' skills for post-hoc verification. This threshold is lower than our clustering threshold because when sampling negative skills, we do not want to admit paraphrased versions of the same skill, where as in our clustering, we only wish to ensure extremely high similarity within each slice, since each slice ultimately only takes on one name. Figure \ref{fig:consistency} shows the inter-annotator agreement, both across different skill annotators (bottom) and verifiers. For verifiers, agreement is very high ($>0.9$) across the board, while agreement is slightly lower for skill annotators. We believe this may relate to the fact that our skill discovery procedure may not result in a complete list of skills, which we note in section \ref{app-sec:limitations}. We argue precision (skill relevance) is more important than recall (completeness), as omitted skills do not affect the veracity of skill-slice accuracy (so long as they are omitted randomly), while incorrect skills would corrupt the signal. 

\section{Details on Routing Proof of Concept}
\label{app-sec:routing}

\begin{table}[]
    \centering
\resizebox{\textwidth}{!}{%
\begin{tabular}{lcccccccccccc}
\toprule
Dataset & SEEDBench & MMLU Pro & MMBench & MMTBench & MME & MMC & MathVista & MMMU & RealWorld-QA & MMVP & VibeEval & MMVet \\
(\# of instances) & (14232) & (12032) & (4329) & (3123) & (2374) & (2126) & (1000) & (900) & (765) & (292) & (269) & (218)\\
Model &  &  &  &  &  &  &  &  &  &  &  &  \\
\midrule
\GeminiPro & 69.94 & 59.47 & 84.18 & 62.36 & 78.64 & 75.40 & 56.10 & 56.33 & 68.10 & 74.00 & 36.43 & 60.55 \\
\ClaudeSonnet & 69.43 & 58.49 & 84.50 & 65.17 & 84.33 & 82.27 & 48.60 & 58.67 & 71.76 & 65.00 & 44.98 & 59.63 \\
GPT-4o & 70.62 & 56.09 & 87.46 & 64.52 & 82.73 & 77.66 & 50.70 & 60.44 & 75.95 & 85.62 & 49.81 & 62.84 \\ \midrule
Random choice & 69.99 & 58.02 & 85.38 & 64.03 & 81.90 & 78.44 & 51.80 & 58.48 & 71.94 & 74.78 & 43.74 & 61.01 \\
Skill Routing (ours) & 71.81 & 62.94 & 87.13 & 65.42 & 85.89 & 82.13 & 55.50 & 58.89 & 76.34 & 80.82 & 49.07 & 59.63 \\
\bottomrule
\end{tabular}
}
    \caption{Per-dataset accuracies for each of the three foundation models we study, as well as accuracies achieved when randomly selecting a model per instance or when routing each instance to the model strongest on the relevant skills.}
    \label{tab:routing_table}
\end{table}%

We route each instance independently, based on skill-slice accuracies computed over the remaining corpus (i.e. everything except for the current instance). For an instance, we first take all of its listed skills, and then remove any for which we do not have a skill-slice with at least 100 instances on the remaining corpus. If no skills remain, we default to \ClaudeSonnet, which is (barely) the best model over the entire corpus. For the remaining skills, we consolidate accuracies over those slices to obtain a single score per model via a weighted average. Namely, the weight for each skill is the inverse of the number of instances in that skill-slice. This way, more specific (i.e. lower frequency) skills for the instance carry higher weight. Then, the model selected for the instance is the one with the highest score (i.e. weighted average of relevant skill-slice accuracies).

This routing protocol is very simple and does not account for numerous sources of variance, such as different levels of difficulty across instances for the same skill, and how skills can be harder when applied in the context of other skills. Nonetheless, we see accuracy improvement using our simple scheme, suggesting that skill-slice insights generalize.

To provide a full breakdown of the effectiveness of our simple approach, we present the per-dataset results in table \ref{tab:routing_table}. We show accuracy per dataset for each of the three foundation models in our study. Also, we show the accuracy obtained with skill routing (our method) or by selecting a model at random -- in the latter case, we show the expected accuracy. For $11$ out of $12$ datasets, skill routing yields higher accuracy than random model selection, with the sole exception occurring for the smallest dataset in our study (where an accuracy measure may not be as reliable due to the smaller sample size). The accuracy gains exceed $3\%$ for $7$ of the $12$ benchmarks studied. Compared to the best model over our entire suite (\ClaudeSonnet), skill routing improves accuracy for $10$ out of $12$ benchmarks, with a tie in one benchmark. GPT-4o, however, beats skill-routing for $5$ benchmarks, though three of these are the smallest benchmarks in our study. 

Overall, we find that our simple approach to skill routing offers accuracy gains under many settings. We interpret this as positive signal in support of the generalizability of skill-slice insights, as well as a direct potential application of our analysis to improve performance. As mentioned above, the simplicity of our scheme may limit its effectiveness. Potential avenues to improve this approach include considering accuracy over slices formed by \emph{multiple} skills used together, or perhaps by training a lightweight router to predict the best model given skill annotations (akin to an interpretable mixture of experts). We leave these explorations to future work.

\section{Details on Probe Consistency Analysis}
\label{app-sec:probing}

We now provide complete details for our probe consistency analysis. 

\subsection{Generating Probing Questions and Checking Consistency }
\label{app-sec:probing-prompts}

To generate probing questions and check consistency, we require multiple calls to an LLM -- we use GPT-4o. Importantly, neither task requires a multimodal LLM, thanks to skill localization.

First, we rephrase a claim linked to a skill of interest with the below prompt. Notice that we form probing questions in a few slightly varied way, including two yes or no questions testing contradictory questions. Qualitatively, this increases the chance that a model contradicts itself when it is unsure, as opposed to giving the same exact incorrect answer (which our consistency analysis cannot detect). We omit prompts for answering probe questions and checking consistency, as these are straightforward and can be viewed in our code release.

\begin{center}
\resizebox{0.75\textwidth}{!}{
\begin{tcolorbox}[colback=blue!1!white, colframe=blue!75!black, title=Probing question generation]
I will provide a specific skill and an excerpt from a step-by-step solution to some problem. Isolate the central claim that best pinpoints the given skill (next to \textit{SKILL TO PINPOINT}). Then, state that claim and generate three questions that can be used to verify the claim, using the following strategy:
\begin{itemize}
    \item Q1 should repeat the claim as a yes or no question.
    \item Q2 should create a close but contradictory claim, and ask that as a yes or no question. \textit{Do not use negations}.
    \item Q3 should be open-ended, whose answer is contained in the claim alone.
\end{itemize}
ANSWER IN THE FORM OF A PYTHON DICTIONARY, as shown in the following example.
\\
\\
\textbf{Example Response Structure}: \\
\textit{SKILL TO PINPOINT}: \textbf{identifying the longest bar}

\textit{EXCERPT}:
Step 1: Identify the Top Companies from the Image   \begin{itemize}
       \item \textbf{Skill:} Perception: Visual Recognition, Bar chart analysis, Identifying the longest bars
       \item \textbf{Evidence:} In the chart, the two companies with the longest bars are AG Insurance and AXA.
       \item \textbf{Conclusion:} AG Insurance and AXA are the companies with the highest market shares.
   \end{itemize}

\textit{CLAIM AND PROBE QUESTIONS}:\\
\texttt{
``` \\
\{\\
    "CLAIM": "The two companies with the longest bars in the chart are AG Insurance and AXA.", \\
    "Q1": "Are the two companies with the longest bars in the chart AG Insurance and AXA?", \\
    "Q2": "Are the two companies with the longest bars in the chart Allianz and AG Insurance?", \\
    "Q3": "Which companies have the longest bars in the chart?" \\
\}\\
```
}
\end{tcolorbox}
}
\end{center}

\subsection{Correlation of Probe Inconsistency and Skill-slice Accuracy}
\label{app-sec:probing-corrs}

\begin{figure}
    \centering
    \includegraphics[width=\linewidth]{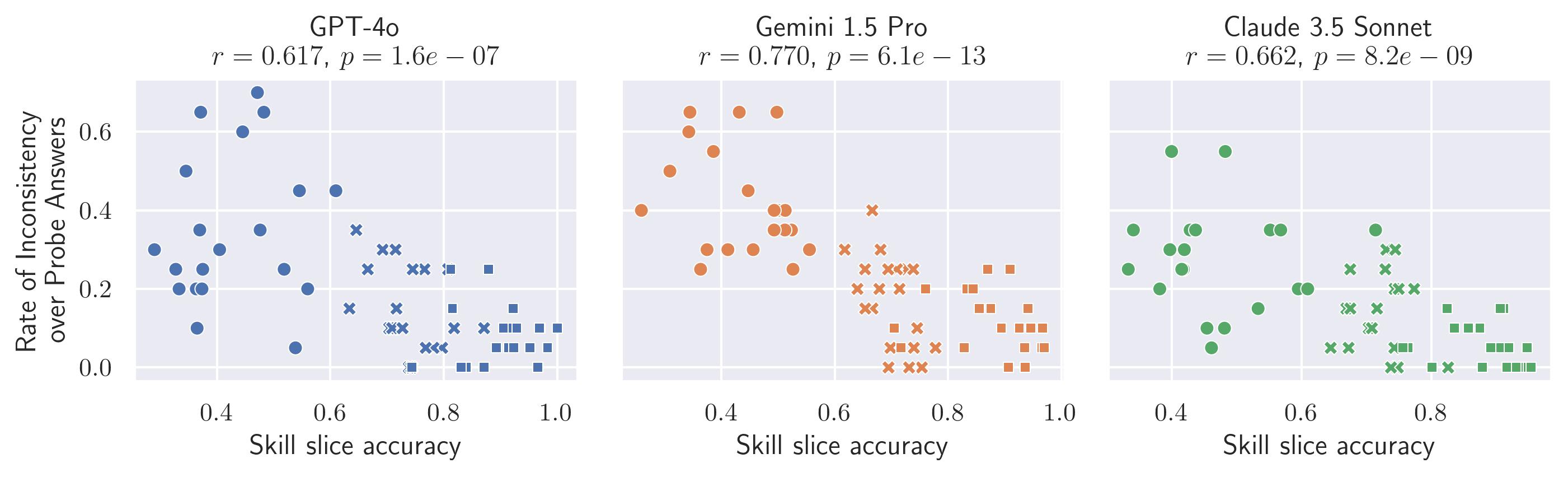}
    \includegraphics[width=0.6\linewidth, trim=0 20 0 20, clip]{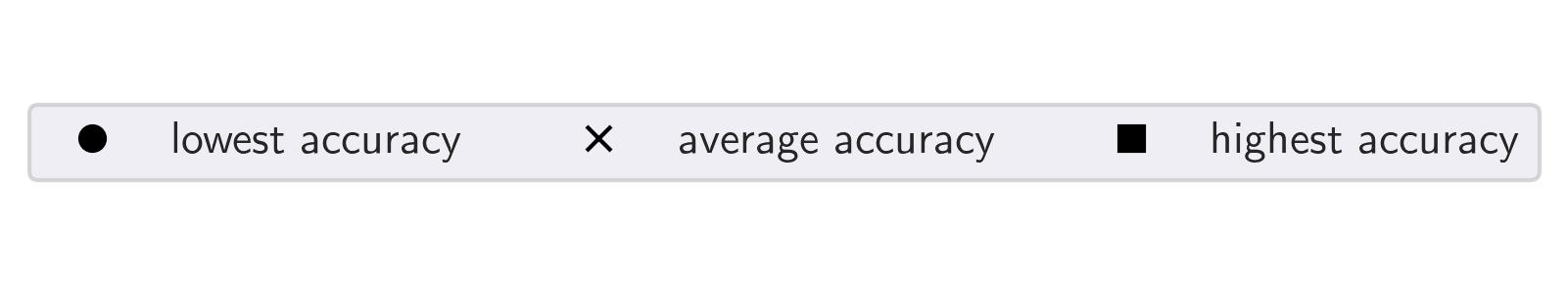}
    \caption{Rate of inconsistency over probe answers vs. slice accuracy per model per skill, as portrayed as a violin plot in Figure \ref{fig:consistency}. For all models, inconsistency negatively correlated with slice accuracy, with an overall Pearson's $r=-0.675$.}
    \label{fig:scatter_consistency}
\end{figure}

\looseness=-1
In figure \ref{fig:scatter_consistency}, we plot probe inconsistency rate vs slice accuracy directly and for each model separately, as opposed to the consolidated violin plot we show in figure \ref{fig:probe-consistency}. We observe reasonably high correlations for all three models, especially given that probe inconsistency rate can only be a factor of $0.05$, as our implementation checks $20$ claims per probed skill.

\begin{figure}
    \centering
    \includegraphics[width=0.75\linewidth]{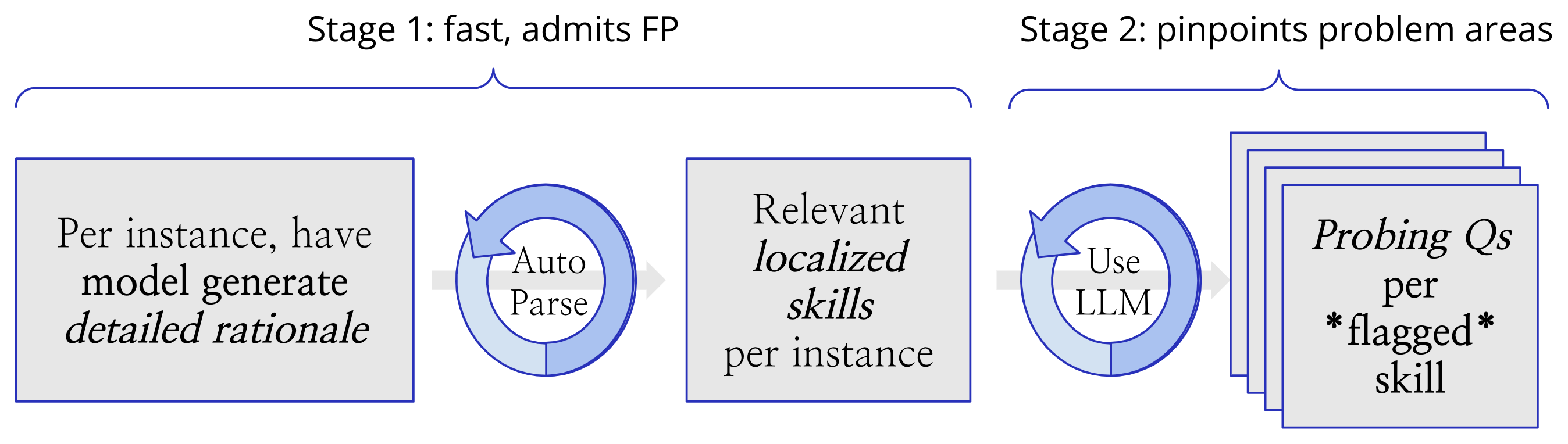}
    \caption{Potential framework for automatic diagnosis of skill deficiencies, leveraging the complementary nature of skill-slice analysis and probe inconsistency.}
    \label{fig:pipeline}
\end{figure}

\subsection{Combining Slice and Probing Analyses Towards Automatic Diagnosis of Skill Deficiencies}
\label{app-sec:probing-e2e}

Lastly, we briefly discuss an end-to-end pipeline for automatically diagnosing skill deficiencies. As shown in figure \ref{fig:pipeline}, a two-stage pipeline can be engineered that first flags skills with low slice accuracy, and then probes those skills. These two stages are complementary in two ways: first, they admit different kinds of errors. If we interpret a deficient skill as a `positive', low skill-slice accuracy can admit false positives due to the presence of a highly co-occurring deficient skill. However, a deficient skill will necessarily have low skill-slice accuracy. Probe inconsistency can help remove these false positives, as a model will always answer consistently (i.e. correctly) to probe questions if the model is proficient at a skill. 

The order of these components is also motivated by the fact that probing requires many more LLM calls, and as such is more expensive than skill-slice analysis. Nonetheless, if cost is not a factor, this procedure could potentially uncover skill-level deficiencies at scale. 

\section{Details on Skill-based Retrieval}
\label{app-sec:retrieval}

We use the same text encoder (\texttt{SFR-Embedding-2\_R}) for any text features, including embedding skills and attributes. We use CLIP ViT-L/14 \cite{clip} for image features. To generate attributes, we prompt GPT-4o to ``List attributes describing the content of given question. **Do not list skills needed to solve the question**, just attributes of the content''. We additionally instruct GPT-4o to avoid answering the question we wish for it to attribute.

\section{Prompt used to generate model outputs}

We note that we employ a simple prompt in obtaining model outputs on the benchmarks we analyze. Namely, we ask models to ``Be concise. Write 'ANSWER: ' followed by your answer. If multiple choices are given, only provide the correct letter.'' As is well documented, accuracies can be improved by engineering stronger prompts, including with model-specific strategies. We opt to choose a single standard prompt for all models, so to avoid giving any single model an advantage due to unequal amounts of time spent on prompt optimization. We leave investigation of these for future work and study models in a simple setting for now.


\end{document}